\documentclass{article} 
\usepackage{iclr2027_preprint,times}

\usepackage{amsmath,amsfonts,bm}

\def\eqref#1{equation~\ref{#1}}

\def\1{\bm{1}}

\def\va{{\bm{a}}}
\def\vb{{\bm{b}}}

\def\vg{{\bm{g}}}
\def\vh{{\bm{h}}}

\def\vk{{\bm{k}}}

\def\vo{{\bm{o}}}

\def\vq{{\bm{q}}}

\def\vu{{\bm{u}}}

\def\vz{{\bm{z}}}

\DeclareMathAlphabet{\mathsfit}{\encodingdefault}{\sfdefault}{m}{sl}
\SetMathAlphabet{\mathsfit}{bold}{\encodingdefault}{\sfdefault}{bx}{n}

\newcommand{\R}{\mathbb{R}}

\newcommand{\softmax}{\mathrm{softmax}}

\usepackage[hidelinks]{hyperref}
\hypersetup{pdftitle={Low-Rank Attention Residuals},pdfauthor={Jonathan Su}}
\makeatletter
\let\preprint@linkurl\hyper@linkurl
\newcommand{\releaselink}[2]{\begingroup\let\hyper@linkurl\preprint@linkurl\href{#1}{\textcolor{blue}{#2}}\endgroup}
\def\hyper@linkurl#1#2{#1\Hy@xspace@end}
\def\hyper@linkfile#1#2#3{#1\Hy@xspace@end}
\def\@hyper@launch run:#1\\#2#3{#2\Hy@xspace@end}
\makeatother
\usepackage{url}
\usepackage{booktabs}
\usepackage{amsmath}
\usepackage{amssymb}
\usepackage{bm}
\usepackage{microtype}
\usepackage{graphicx}
\usepackage{float}
\usepackage{placeins}
\usepackage{xcolor}
\usepackage{enumitem}
\usepackage{pgfplots}
\usepackage{listings}
\pgfplotsset{compat=1.18}

\definecolor{pycharmBackground}{HTML}{141414}
\definecolor{pycharmText}{HTML}{E8E8E8}
\definecolor{pycharmKeyword}{HTML}{FF8A3D}
\definecolor{pycharmFunction}{HTML}{FFD54F}
\definecolor{pycharmBuiltin}{HTML}{C586C0}
\definecolor{pycharmArgument}{HTML}{E0A56A}
\definecolor{pycharmNumber}{HTML}{3FD5FF}
\definecolor{pycharmString}{HTML}{4DC742}
\definecolor{pycharmComment}{HTML}{79C8E8}
\definecolor{pycharmLineNumber}{HTML}{999999}

\lstdefinestyle{lrattnres}{
    language=Python,
    basicstyle=\fontencoding{T1}\fontfamily{fvm}\fontsize{8.5}{10}\selectfont\color{pycharmText},
    backgroundcolor=\color{pycharmBackground},
    keywordstyle=\bfseries\color{pycharmKeyword},
    deletekeywords={None},
    keywordstyle=[2]\color{pycharmBuiltin},
    deletekeywords=[2]{sum},
    morekeywords=[2]{len,min,range,enumerate,zip},
    keywordstyle=[3]\color{pycharmArgument},
    morekeywords=[3]{dim,start},
    keywordstyle=[4]\color{pycharmFunction},
    morekeywords=[4]{route,lr_attnres,rms_norm},
    keywordstyle=[5]\color{pycharmBuiltin},
    morekeywords=[5]{None},
    identifierstyle=\color{pycharmText},
    stringstyle=\bfseries\color{pycharmString},
    commentstyle=\color{pycharmComment},
    literate=*{-}{{{\char45}}}1
              {0}{{{\color{pycharmNumber}0}}}1
              {1}{{{\color{pycharmNumber}1}}}1
              {2}{{{\color{pycharmNumber}2}}}1
              {3}{{{\color{pycharmNumber}3}}}1
              {4}{{{\color{pycharmNumber}4}}}1
              {5}{{{\color{pycharmNumber}5}}}1
              {6}{{{\color{pycharmNumber}6}}}1
              {7}{{{\color{pycharmNumber}7}}}1
              {8}{{{\color{pycharmNumber}8}}}1
              {9}{{{\color{pycharmNumber}9}}}1,
    numbers=left,
    numberstyle=\fontencoding{T1}\fontfamily{fvm}\fontsize{6.5}{8}\selectfont\color{pycharmLineNumber},
    numbersep=8pt,
    xleftmargin=2em,
    xrightmargin=0.5em,
    framexleftmargin=1.5em,
    columns=fullflexible,
    keepspaces=true,
    showstringspaces=false,
    upquote=true,
    frame=single,
    rulecolor=\color{pycharmBackground},
    framerule=0.4pt,
    framesep=5pt,
    aboveskip=0pt,
    belowskip=0pt
}

\title{Low-Rank Attention Residuals}

\author{Jonathan Su \\
Independent Researcher \quad \texttt{270985@learning.gsis.edu.hk}
}

\usepackage{xspace}

\newcommand{\AttnRes}{\textsc{AttnRes}\xspace}
\newcommand{\BlockAttnRes}{\textsc{Block AttnRes}\xspace}
\newcommand{\LRAttnRes}{\textsc{LR-AttnRes}\xspace}
\newcommand{\fastattnres}{\textsc{Fast-AttnRes}\xspace}
\newcommand{\PLRAttnRes}{\textsc{Projected LR-AttnRes}\xspace}

\providecommand{\R}{\mathbb{R}}
\providecommand{\softmax}{\operatorname{softmax}}
\providecommand{\rmsnorm}{\operatorname{RMSNorm}}
\providecommand{\tailr}{\operatorname{tail}_r}
\providecommand{\vo}{\mathbf{o}}
\providecommand{\vh}{\mathbf{h}}
\providecommand{\vk}{\mathbf{k}}
\providecommand{\vq}{\mathbf{q}}
\providecommand{\va}{\mathbf{a}}
\providecommand{\vz}{\mathbf{z}}
\providecommand{\vb}{\mathbf{b}}
\providecommand{\vu}{\mathbf{u}}
\providecommand{\vg}{\mathbf{g}}
\providecommand{\valpha}{\boldsymbol{\alpha}}

\begin{document}

\maketitle

\begin{abstract}
Attention Residuals (\AttnRes) replace the fixed residual sum with depth-wise attention over previous sub-layer outputs in Large Language Models (LLMs), but use each output as both a full-dimensional key and value. This couples routing with representation and makes the cost of computing depth-routing scores scale with hidden width $d$. We propose Low-Rank Attention Residuals (\LRAttnRes), which keep full-dimensional residual values while using $r$-dimensional keys, with $r < d$, for routing. \LRAttnRes uses the last $r$ dimensions of each value as the routing key, reducing total residual-side FLOPs while still improving performance. Comprehensive sweeps across the number of blocks ($N$) and $r$ show that depth-wise routing can be effective with far fewer dimensions than the model width. At both $1$B and $4$B parameters with $r = d/4$, \LRAttnRes achieves lower final validation loss, higher average downstream accuracy, and higher measured training-step throughput than standard \AttnRes. We also provide a fused kernel supporting standard and low-rank routing. We release all \releaselink{https://github.com/jon123boss/LR-AttnRes}{code}, the \releaselink{https://github.com/jon123boss/fast-attnres}{kernel}, and all trained \releaselink{https://huggingface.co/Jonnester}{models} to facilitate future research.
\end{abstract}

\begin{figure}[!h]
    \centering
    \includegraphics[width=\linewidth]{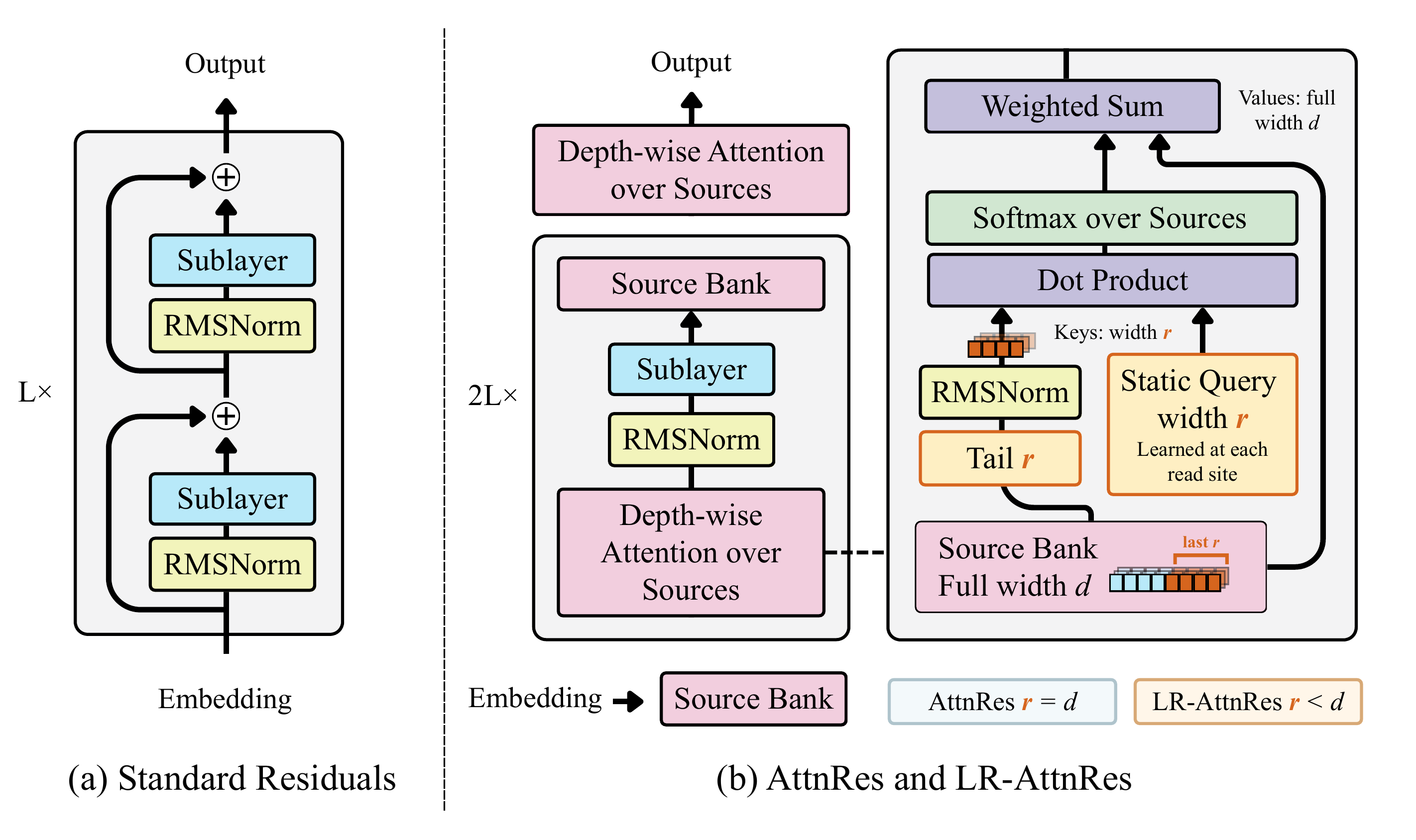}
    \caption{
    Residual and routing paths. (a) Standard PreNorm residuals. (b) Shared architecture for \AttnRes{} ($r=d$) and \LRAttnRes{} ($r<d$). In full routing, the source bank retains the embedding and appends each sub-layer output. Keys use the last $r$ coordinates of each source before RMSNorm (values retain all $d$ coordinates). Queries are learned, input-independent, and specific to each read site. Softmax is over sources. The $L$ Transformer blocks contain $2L$ sub-layers, with an additional depth-wise attention read before the output head. During block routing, the current block's partial sum is included as an additional source. Once the block is complete, its sum is retained as a separate source in the source bank, while a new partial sum accumulates for the next block.
    }
    \label{fig:lr_attnres_architecture}
\end{figure}

\clearpage
\section{Introduction}
\begingroup

Residual connections are central to the optimization of deep networks~\citep{resnet}. This is especially true in Transformer-based large language models~\citep{attnisallyouneed}, where residual streams allow information to accumulate across depth and also serve as gradient highways for optimization. In a PreNorm decoder~\citep{prenorm}, each attention or feed-forward sub-layer reads a normalized version of the current stream and writes an output back into it. This fixed additive rule is simple and stable, but it gives every previous output a fixed unit coefficient. The model can learn what each sub-layer writes, yet it cannot directly choose which previous writes should be emphasized when constructing the next hidden state. \textbf{Attention Residuals} (\textbf{\AttnRes}) address this limitation by replacing fixed accumulation with attention over depth~\citep{kimiteam2026attentionresiduals}.

At each residual site, the hidden state is formed as a learned weighted combination of earlier sub-layer outputs. This is analogous to token attention, but the ``tokens'' are previous sub-layer outputs. The full version attends over all previous sub-layer outputs, while \textbf{\BlockAttnRes} groups several outputs into block summaries to reduce storage and communication.

The standard \AttnRes{} design makes each sub-layer output
$\vo_i \in \R^d$ serve two roles simultaneously: it is both the
value mixed into the residual stream and, after normalization, the
key used to determine how strongly that source should be selected.
These roles need not require the same representation. A residual
output should remain a full-capacity carrier of information for
downstream layers, whereas a routing key only needs to distinguish
among a small set of depth sources. 

This disparity is striking at production scale. Kimi K3 ($d=7168$) groups its 93 layers into $8$ \AttnRes{} blocks, using full-width queries and keys to select among at most $9$ sources per depth read, including the embedding~\citep{kimik3}. Its token context extends to $10^6$ positions, roughly five orders of magnitude more candidates. Although depth-wise and sequence-wise attention solve different problems, source cardinality alone offers little justification for full-width depth routing, and the observed query concentration suggests that narrower routing representations may suffice.

Across Kimi K3's 186 active \AttnRes{} queries, folding the RMSNorm affine weights into the query projections gives a mean Shannon coordinate effective dimension of $602.6$, just $8.4\%$ of their width. Moreover, $45.5\%$ of folded-query coordinates are exactly zero under the released implementation's FP32 arithmetic, almost entirely because of underflow during folding. Appendix~\ref{app:kimi_query_utilization} gives the definition and layerwise results. These observations motivate separating routing-key width from value width.

We introduce \textbf{Low-Rank Attention Residuals} (\textbf{\LRAttnRes}), which retain full-dimensional residual values while computing depth-routing scores with $r$-dimensional keys, where $r < d$. \LRAttnRes uses the final $r$ value dimensions as keys, yielding a net reduction in added FLOPs compared to \AttnRes. Section~\ref{sec:depth_vs_token_dimensions} gives conditions for converting \AttnRes{} checkpoints to full-width-RMS \LRAttnRes{} with identical outputs in exact arithmetic, with potential applicability to Kimi K3. Figure~\ref{fig:lr_attnres_architecture} illustrates the construction. We also explore \textbf{Projected LR-AttnRes}, which generates exogenous routing keys through auxiliary projections. This alternative is discussed in Appendix~\ref{app:projected_lr}.

We evaluate $32$ configurations at $0.5$B parameters and width $d=1024$, spanning full routing, block counts $N\in\{4,8,16\}$, and ranks $r\in\{16,32,64,128,256,512,768,1024\}$. The best configuration uses $N=8,r=128=d/8$. At $1$B and $4$B, \LRAttnRes{} with $r=d/4$ achieves lower final validation loss than \AttnRes{} and increases average accuracy across eight downstream benchmarks by $0.40$ and $0.57$ percentage points, respectively.

\textbf{Our contributions are:}
\begin{itemize}
    \item We introduce \LRAttnRes{}, a depth-routing mechanism that retains full-dimensional values while reducing the routing-key width from $d$ to $r < d$.
    \item We develop \textbf{\fastattnres}{}, a fused kernel supporting both standard \AttnRes{} and \LRAttnRes{} (Appendix~\ref{app:fast_attnres}).
    \item We characterize static-query expressivity under full-width normalization and local routing feedback under both normalization choices, and evaluate rank, grouping, seed replication, and normalization effects.
\end{itemize}

\endgroup

\clearpage
\section{Related Work}
\label{sec:related_work}

A broad line of work modifies the residual pathway. DenseNet concatenates earlier outputs, while DenseFormer combines layer
outputs with learned static depth-wise weights~(\citealp{densenet} and \citealp{denseformer}).
Other methods introduce more expressive cross-layer access, including retrospective layer attention, learned augmented residual layers, multi-stream or hyper-connection mechanisms, and input-dependent depth mixing~(\citealp{DBLP:conf/iclr/FangCC0TL23}, \citealp{laurel}, \citealp{hyperconnections}, \citealp{mhc}, and \citealp{muddformer}).

\AttnRes is the most direct predecessor of our work~\citep{kimiteam2026attentionresiduals}. In standard
\AttnRes, however, the same full-dimensional tensor is used both as the residual value and as the routing key. \LRAttnRes keeps the cross-layer routing framework but isolates a more specific design axis: the representation and dimensionality
used to score depth sources. We show that low-dimensional routing keys can decrease validation loss while reducing FLOPs.

\section{Method}
\label{sec:method}

\subsection{Depth residual notation}

Consider a decoder-only Transformer with $L$ blocks and $M=2L$ residual-writing sub-layers: one attention sub-layer and one feed-forward sub-layer per block. For clarity, equations omit batch and sequence dimensions. Let $\vo_i \in \R^d$ denote the output written by sub-layer $i$, and let $\vo_0 \in \R^d$ denote the token embedding. We index residual read sites by $t\in\{1,\ldots,M+1\}$, where site $t$ reads from sources written before that site. A standard residual stream supplies the fixed sum
\begin{equation}
    \vh_t = \vo_0 + \sum_{i=1}^{t-1}\vo_i
    \label{eq:standard_residual}
\end{equation}
to residual site $t$.

\AttnRes replaces this fixed accumulation with learned attention over depth. At residual site $t$, the hidden state is formed as a weighted mixture over a source set $\mathcal{S}_t \subseteq \{0,\ldots,t-1\}$:
\begin{align}
    \alpha_{t,i}
    &=
    \softmax_{i\in\mathcal{S}_t}
    \left(
        \vq_t^\top \rmsnorm(\vo_i)
    \right),
    \label{eq:attnres_weight}\\
    \vh_t
    &=
    \sum_{i\in\mathcal{S}_t}\alpha_{t,i}\vo_i .
    \label{eq:attnres_output}
\end{align}
Here $\vq_t\in\R^d$ is a learned pseudo-query associated with residual read site $t$. Thus, unlike a standard residual stream, the model can learn which previous sub-layer outputs to read from rather than always summing all previous outputs equally. The same tensor $\vo_i$ is used as the value in Eq.~\ref{eq:attnres_output} and, after normalization, as the key in Eq.~\ref{eq:attnres_weight}.

\paragraph{Full \AttnRes.}
The full version attends to every previous residual-writing source:
\begin{equation}
    \mathcal{S}_t^{\mathrm{full}}
    =
    \{0,1,\ldots,t-1\}.
    \label{eq:full_attnres_source_set}
\end{equation}
Full \AttnRes gives the maximum depth-routing flexibility, because every residual site can directly select from the embedding and all earlier attention and feed-forward outputs. The cost is that all previous full-dimensional outputs must be retained as attention sources.

\paragraph{\BlockAttnRes.}
\BlockAttnRes reduces the number of stored sources by grouping contiguous sub-layer outputs into block summaries. At a residual site $t$, the model attends to the embedding, all completed block summaries, and the current partial block if one exists. The partial block contains the outputs that have been written since the most recent completed block boundary but have not yet accumulated enough outputs to form a completed block.

Let $B$ be a completed block containing $c=|B|$ sub-layer outputs. The block summary is the sum
\begin{equation}
    \vb_B^{\mathrm{sum}}
    =
    \sum_{i\in B}\vo_i .
    \label{eq:block_attnres_sum}
\end{equation}
Now consider a residual site $t$ that lies inside an unfinished block. Let $P_t \subseteq \{1,\ldots,t-1\}$ denote the current partial block: the set of residual-writing sub-layers after the last completed block boundary and before site $t$. If $P_t$ is nonempty, its partial summary is
\begin{equation}
    \vb_{P_t}^{\mathrm{sum}}
    =
    \sum_{i\in P_t}\vo_i .
    \label{eq:block_attnres_partial_sum}
\end{equation}

\subsection{Low-rank routing with full-dimensional values}

\LRAttnRes keeps the same full-dimensional value path as \AttnRes but computes depth-routing scores in a smaller key space. Each source is represented as
\begin{equation}
    \text{source } i:
    \qquad
    (\vo_i,\vk_i),
    \qquad
    \vo_i\in\R^d,
    \quad
    \vk_i\in\R^r,
    \quad
    r< d.
    \label{eq:lr_source_pair}
\end{equation}
The value $\vo_i$ remains the ordinary full-dimensional sub-layer output. The key $\vk_i$ is an $r$-dimensional routing descriptor used only to select among depth sources. At residual site $t$, a learned static query $\vq_t\in\R^r$ scores normalized keys:
\begin{align}
    \bar{\vk}_i
    &=
    \rmsnorm(\vk_i),
    \label{eq:lr_key_norm}\\
    \alpha_{t,i}
    &=
    \softmax_{i\in\mathcal{S}_t}
    \left(
        \vq_t^\top \bar{\vk}_i
    \right),
    \label{eq:lr_attnres_weight}\\
    \vh_t
    &=
    \sum_{i\in\mathcal{S}_t}\alpha_{t,i}\vo_i .
    \label{eq:lr_attnres_output}
\end{align}

This changes only the routing descriptor. The output mixed into the residual stream is still the full $d$-dimensional $\vo_i$, so \LRAttnRes does not impose a low-rank bottleneck on residual values. \LRAttnRes obtains $\vk_i$ by slicing the value, as described below.

\subsection{Routing keys}
\label{sec:sliced_lr_keys}

\LRAttnRes uses the last $r$ dimensions of the residual value as the key, without a learned key-projection path:
\begin{equation}
    \vk_i^{\mathrm{S}}
    =
    \tailr(\vo_i)
    =
    (\vo_i)_{d-r+1:d}
    \in \R^r,
    \qquad
    \bar{\vk}_i^{\mathrm{S}}
    =
    \rmsnorm(\vk_i^{\mathrm{S}}).
    \label{eq:sliced_key}
\end{equation}
The embedding source uses the same rule:
\begin{equation}
    \vo_0 = \operatorname{Embed}(x),
    \qquad
    \vk_0^{\mathrm{S}}
    =
    \tailr(\vo_0),
    \qquad
    \bar{\vk}_0^{\mathrm{S}}
    =
    \rmsnorm(\vk_0^{\mathrm{S}}).
    \label{eq:sliced_embedding_key}
\end{equation}

The last $r$ dimensions are a convention: any fixed subset is equivalent up to a global permutation of hidden coordinates. Let $P_r$ select these coordinates. For a sub-layer output $\vo_i=W_i^O\va_i$ with projection input $\va_i$, the key $P_r\vo_i=(P_rW_i^O)\va_i$ is learned through the selected rows of the value projection, without an auxiliary projection. It can be implemented as a view into the cached value. When $r=d$, \LRAttnRes reduces to \AttnRes.

\paragraph{Partial-block routing.}
Each sub-layer output updates the full-width partial sum $\vb\leftarrow\vb+\vo_i$. Its normalized key is $\rmsnorm(\tailr(\vb))$: accumulation precedes slicing and normalization. At a block boundary, the partial sum becomes a completed source and the partial accumulator resets to empty.

Figure~\ref{fig:lr_attnres_pseudocode} in Appendix~\ref{app:lr_attnres_pseudocode} summarizes full and block routing for both \AttnRes{} ($r=d$) and \LRAttnRes{} ($r<d$) in a shared implementation using PyTorch pseudocode.

\subsection{Efficient inference compatibility}
\label{sec:efficient_inference_compatibility}

\LRAttnRes preserves the conditions required by the efficient inference strategy used by base \AttnRes. The two-phase strategy for \BlockAttnRes relies on two properties. First, residual-site queries are input-independent. Second, the inter-block and intra-block attention results can be merged with an online-softmax update. \LRAttnRes preserves both properties. It changes only the key representation used for routing, while keeping the residual values in $\mathbb{R}^d$.

\clearpage
\section{Experiments}
\label{sec:experiments}

\subsection{Experimental setup}
\label{sec:experimental_setup}

\paragraph{Architecture.}
The $0.5\text{B}$ sweep uses a 24-layer decoder-only Transformer with hidden dimension $d=1024$, 16 attention heads, SwiGLU hidden width $m=2816$, RoPE~\citep{roformer} base $\theta=500{,}000$, QK normalization~\citep{qknorm}, PreNorm, FlashAttention~\citep{flashattn}, untied input and output embeddings, and a vocabulary size of $100{,}277$. The context length is 2048 tokens. Each Transformer block has one attention sub-layer and one MLP sub-layer, giving $2L=48$ residual writers. Depth-wise pseudo-queries are initialized to zero and key normalization is enabled. At $1\text{B}$ and $4\text{B}$, paired standard \AttnRes{} and \LRAttnRes{} models use $N=8$, matching Kimi K3~\citep{kimik3}. We fix $r/d=1/4$ at both scales to test transfer without rank tuning: this reduces key width by $75\%$ while retaining a larger fraction of coordinates than the best-performing $0.5\text{B}$ ratio, $r/d=1/8$ (Section~\ref{sec:validation_rank}).

\paragraph{Training.}
The sweep models are trained from scratch on Ultra-FineWeb~\citep{ultrafineweb} for 10B tokens with document masking. All scales use a global batch size of 262{,}144 tokens. We optimize with Muon~\citep{jordan2024muon} and Adam~\citep{adam}. Muon uses learning rate $10^{-3}$, momentum $0.95$, and weight decay $0.1$. Adam uses learning rate $3\times10^{-4}$, $\beta_1=0.9$, $\beta_2=0.95$, and weight decay $0$. We include a z-loss regularizer with coefficient $10^{-5}$. Gradients are clipped to norm 1.0. We use a linear schedule with 2000 warmup steps and $20\%$ warmdown steps. Final validation reports token-averaged cross-entropy on 100M held-out tokens. Two sweep configurations are each trained three times in total (Section~\ref{sec:seed_rms}). The $1\text{B}$ and $4\text{B}$ pairs use $20\text{B}$ training tokens.

\subsection{Routing-key rank and validation loss}
\label{sec:validation_rank}

We evaluate all 32 combinations of eight ranks and four routing settings. Appendix~\ref{app:validation_results} reports all 32 sweep configurations.

\begin{figure}[!ht]
    \centering
    \includegraphics[width=0.83\linewidth]{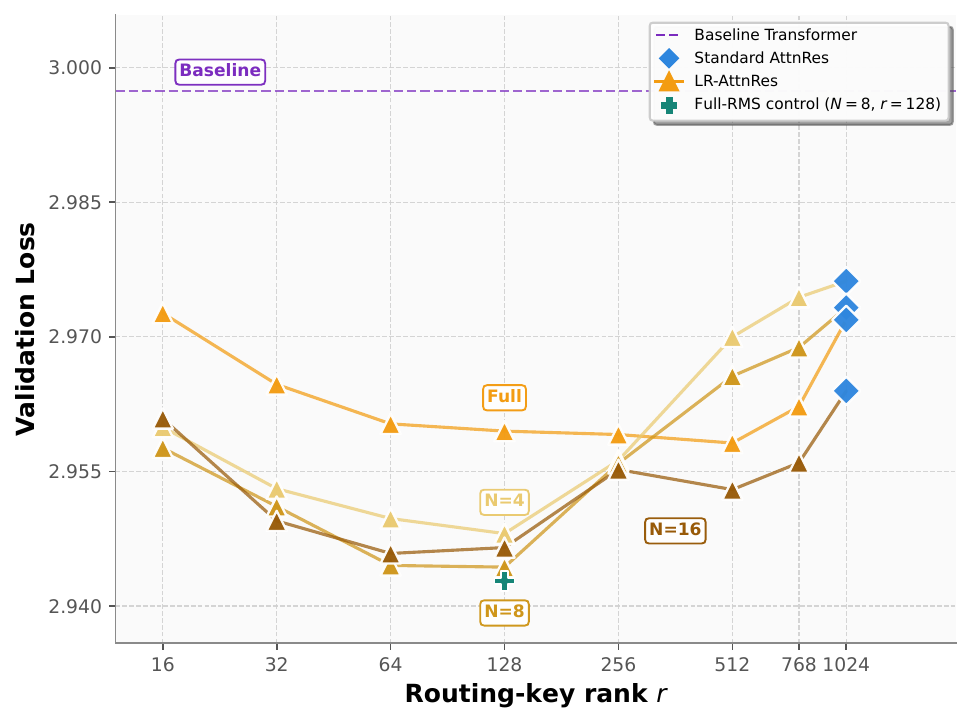}
    \caption{
    Validation loss versus routing-key rank $r$ (base-2 log axis) at $0.5\text{B}$. Triangles denote \LRAttnRes{}. Blue diamonds denote standard \AttnRes{}. Only the $N=8,r=128$ triangle and $N=8,r=d=1024$ diamond plot means across three training seeds; all other sweep points show single runs. The teal plus is the full-width-RMS control. The purple dashed line is the baseline. Appendix~\ref{app:validation_results} gives numerical results.
    }
    \label{fig:validation_vs_rank}
\end{figure}

All 32 routed configurations improve on the baseline loss of $2.997$. 27 of the 28 low-rank configurations improve on their $r=d$ endpoint (full routing at $r=16$ is the exception). The lowest observed loss is $2.944$ at $N=8,r=128$, improving by $0.03074$ over standard $N=8$ \AttnRes{} and by $0.02035$ over the best standard setting, $N=16$. At $N=8,r=64$, loss is only $8.978\times10^{-4}$ higher. These minima are based on a single training run per configuration.

The preferred rank depends on source grouping. Full \LRAttnRes{} improves from $2.973$ at $r=16$ to $2.958$ at $r=512$, then worsens to $2.962$ at $r=768$ and $2.972$ at $r=d$. All three block settings instead attain their minima at $r=64$ or $128$. Neither the smallest nor the largest routing space is uniformly best.

The $N=8,r=128$ configuration also reduces added residual-side FLOPs by approximately $44\%$ relative to standard $N=8$ \AttnRes{}, from $0.1754\%$ to $0.09865\%$ of the non-embedding Transformer core. Appendix~\ref{app:validation_results} presents the complete validation--FLOPs comparison, and Appendix~\ref{app:costs} details the computational accounting.

\FloatBarrier
\subsection{Block granularity}
\label{sec:block_granularity}

Block granularity and rank interact: $N=8$ performs best at $r=16,64,128$, and $N=16$ at the other tested ranks. The minima for $N=4,8,16$ are $2.948$, $2.944$, and $2.946$. Increasing $N$ expands the source set but does not consistently improve validation loss.

At $r=256$, block losses span only $0.001016$, a near-tie that could reflect training-seed variability. At $r=512$, the spread grows to $0.01702$: $N=16$ improves while $N=4,8$ worsen.

The lack of a consistent benefit from larger $N$ may reflect changes in the sources themselves. Changing $N$ changes which outputs are summed together and assigned a shared routing weight. These outputs can reinforce or cancel before normalization, changing the routing keys as well as the number of sources (Appendix~\ref{app:block_routing_geometry}). These results favor joint selection of $N$ and $r$, rather than a universal U-shaped rank response or optimal block count.

\subsection{Seed replication and normalization control}
\label{sec:seed_rms}

We replicate the selected $N=8,r=128$ configuration and standard $N=8$ \AttnRes{} with two additional seeds. Figure~\ref{fig:validation_vs_rank} shows the three-seed means. Mean $\pm$ sample standard deviation is $2.973\pm0.0009852$ for \AttnRes{} and $2.944\pm0.001777$ for \LRAttnRes{}. The mean paired reduction is $0.02893$, with improvements in all three pairs. Per-seed losses are in Appendix~\ref{app:validation_results}.

To test whether slice-based normalization is necessary for the gain, we train a full-width-RMS control. Let $P_r$ select the final $r$ coordinates of a value $\vo_i\in\R^d$. The two constructions are
\begin{equation}
\begin{aligned}
\bar{\vk}_i^{\mathrm{slice}}
&=\rmsnorm(P_r\vo_i)
=\frac{P_r\vo_i}{\sqrt{\|P_r\vo_i\|_2^2/r+\epsilon}},\\[3pt]
\bar{\vk}_i^{\mathrm{control}}
&=P_r\,\rmsnorm(\vo_i)
=\frac{P_r\vo_i}{\sqrt{\|\vo_i\|_2^2/d+\epsilon}}.
\end{aligned}
\label{eq:rms_control}
\end{equation}
The default method slices first, then applies RMSNorm to the $r$ selected coordinates. The control applies RMSNorm to all $d$ coordinates first, then slices. Both score sources with $\vq_t^\top\bar{\vk}_i$ and mix full-width values. The control reaches $2.943$: $0.03157$ below standard \AttnRes{} and $8.345\times10^{-4}$ below slice-normalized \LRAttnRes{}. Thus the gain persists with full-width normalization. Section~\ref{sec:routing_feedback} characterizes the resulting routing feedback, and Section~\ref{sec:depth_vs_token_dimensions} examines expressivity through frozen-checkpoint conversion.

\subsection{Results at 1B and 4B parameters}
\label{sec:preliminary_1b}

\begin{table}[!ht]
    \centering
    \caption{Results after $20\text{B}$ training tokens. Accuracy averages eight downstream tasks.}
    \label{tab:preliminary_1b}
    \vspace{0.65em}
    \small
    \setlength{\tabcolsep}{5pt}
    \newcommand{\MainResultsRows}{%
1B & \AttnRes{} & 41.87 & 2.800 \\
1B & \LRAttnRes{} & \textbf{42.27} & \textbf{2.780} \\
\midrule
4B & \AttnRes{} & 42.97 & 2.738 \\
4B & \LRAttnRes{} & \textbf{43.54} & \textbf{2.706} \\
}
    \begin{tabular}{@{}llrr@{}}
        \toprule
        Model & Method & \shortstack{Avg. accuracy\\(\%) $\uparrow$} & \shortstack{Final val.\\loss $\downarrow$} \\
        \midrule
        \MainResultsRows
        \bottomrule
    \end{tabular}
\end{table}

The $1\text{B}$ models use 24 layers, width $d=1536$, 24 heads, and SwiGLU width $4224$. The $4\text{B}$ models use 44 layers, $d=2560$, 40 heads, and SwiGLU width $7040$. Table~\ref{tab:preliminary_1b} shows validation-loss reductions of $0.01946$ and $0.03204$, with eight-task mean accuracy gains of $0.40$ and $0.57$ percentage points. Appendix~\ref{app:preliminary_1b} gives the evaluation protocol and per-task scores.

\begin{figure}[!ht]
    \centering
    \includegraphics[width=0.88\linewidth]{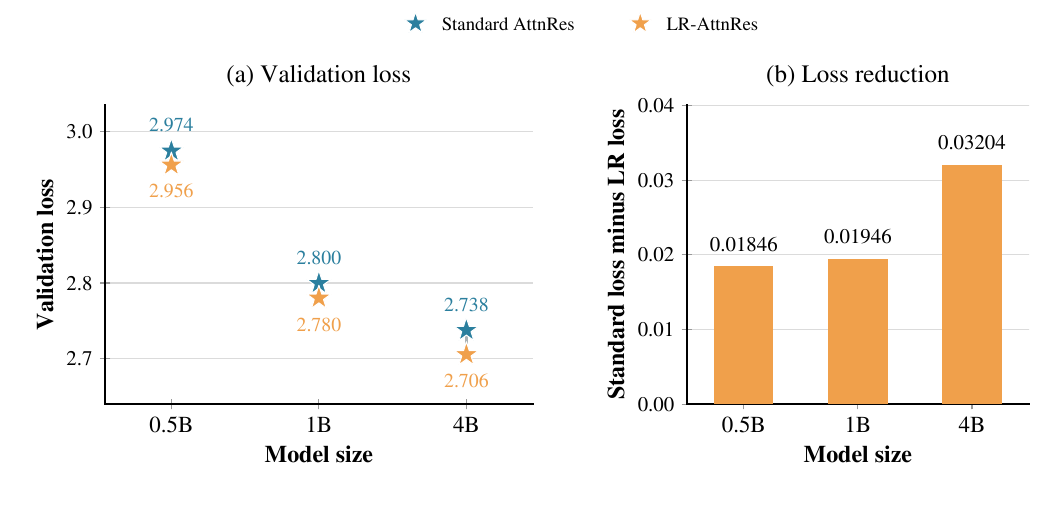}
    \caption{Validation loss across model sizes with block routing $N=8$ and LR rank $r=d/4$ ($r=256,384,640$). Left: final validation loss. Right: standard \AttnRes{} minus \LRAttnRes{} loss. The $0.5\text{B}$ pair is trained for $10\text{B}$ tokens, and the $1\text{B}$ and $4\text{B}$ pairs for $20\text{B}$ tokens.}
    \label{fig:model_size_comparison}
\end{figure}

\FloatBarrier

\subsection{Effective number of sources}
\label{sec:effective_sources}

\begin{figure}[!ht]
    \centering
    \includegraphics[width=0.48\linewidth]{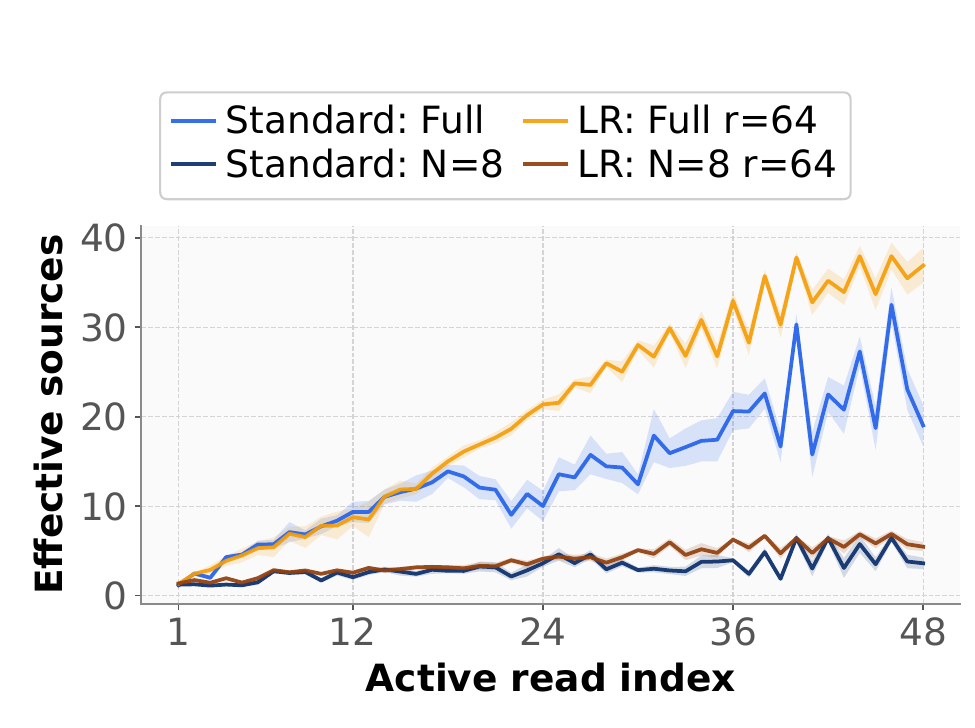}
    \hfill
    \includegraphics[width=0.48\linewidth]{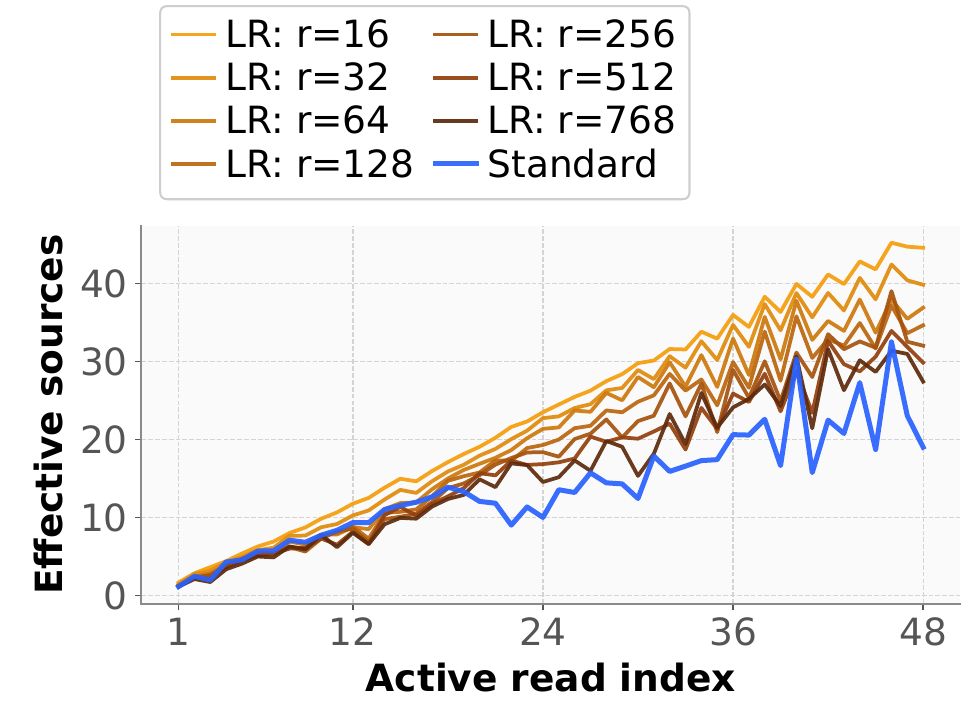}
    \caption{
    Effective number of residual sources across residual read sites. Left: comparison between full and block routing for standard \AttnRes{} (Standard) and \LRAttnRes{} (LR). Right: rank sweep for full \LRAttnRes{}, with standard full \AttnRes{} shown in blue. Lines show token means; shading in the left panel indicates $\pm1$ standard deviation across $1\mathrm{M}$ held-out tokens. Lower values indicate sharper source selection.
    }
    \label{fig:effective_sources}
\end{figure}

We next analyze how many residual sources are effectively used by each depth-wise attention distribution. For a residual read site $t$ with source set $\mathcal{S}_t$, let
$\valpha_t(x)$ denote the source-attention distribution for token position $x$, where $x$ ranges over the batch and sequence positions used in the probe. We define the effective number of sources as the entropy perplexity of this distribution:
\begin{equation}
    N_{\mathrm{eff}}(t,x)
    =
    \exp\left(
        H(\valpha_t(x))
    \right),
    \qquad
    H(\valpha_t(x))
    =
    -\sum_{i\in\mathcal{S}_t}
    \alpha_{t,i}(x)\log \alpha_{t,i}(x).
    \label{eq:effective_sources}
\end{equation}
This equals $1$ for selection of a single source and $k$ for uniform attention over $k$ sources. We average this quantity over 1M held-out tokens at each of 48 query-bearing reads. Diagnostic figures number these active reads 1--48, corresponding to $t=2,\ldots,49$ in Section~3.1. At the final read, replacing standard keys with rank-$64$ keys raises the effective source count from $19.02$ to $36.90$ under full routing and from $3.62$ to $5.47$ under $N=8$ block routing. Within block settings, low-rank keys produce less concentrated attention distributions, spreading weight across more residual sources.

Broader routing is not itself a predictor of lower loss. In the full-routing rank sweep, the final-site effective count decreases from $44.57$ at $r=16$ to $29.85$ at $r=512$, even as validation loss improves from $2.973$ to $2.958$. At $r=768$, the count decreases further to $27.42$. 

\FloatBarrier
\subsection{Routing feedback}
\label{sec:routing_feedback}

Slicing changes how a source influences both its contribution and its selection. For one read, write $z_i$ for its source-local score, $\alpha_i=\softmax_i(z)$, and $\vg=\nabla_{\vh}\mathcal L$. Treating the source slots as independent inputs, the gradient through this read decomposes as
\begin{equation}
\nabla_{\vo_i}\mathcal L
=\underbrace{\alpha_i\vg}_{\text{direct value gradient}}
+\underbrace{\alpha_i\langle\vg,\vo_i-\vh\rangle\nabla_{\vo_i}z_i}_{\text{routing feedback}}.
\label{eq:value_routing_gradient}
\end{equation}
For finite logits, the local source-to-output Jacobian has rank at least $d-1$, irrespective of $r$, so narrow scoring does not impose a rank-$r$ bottleneck on local value transmission.
Let $\Pi_r=P_r^\top P_r$ and $\rho_m(\vu)=\sqrt{\|\vu\|_2^2/m+\epsilon}$. The two normalization choices in Eq.~\ref{eq:rms_control} give
\begin{equation}
\begin{aligned}
(I-\Pi_r)\nabla_{\vo_i}z_i^{\mathrm{slice}}&=0,\\
(I-\Pi_r)\nabla_{\vo_i}z_i^{\mathrm{control}}
&=-\frac{z_i^{\mathrm{control}}}{d\,\rho_d(\vo_i)^2}(I-\Pi_r)\vo_i.
\end{aligned}
\label{eq:routing_feedback_support}
\end{equation}
Thus unselected coordinates receive only the direct value contribution under slice normalization. A perturbation $\vu$ with $P_r\vu=0$ leaves the routing weights unchanged and changes the mixed output by $\alpha_i\vu$. The control retains dependence on the norm of these coordinates, so complete local isolation is not necessary for the gain observed in Section~\ref{sec:seed_rms}. These properties only concern the current read. This local separation does not establish that routing feedback harms learning. Appendix~\ref{app:routing_feedback_proof} gives the derivation. Gradient-norm measurements are reported in Appendices~\ref{app:projected_gradients} and~\ref{app:large_model_gradients}.

\subsection{Static-query span and normalization}
\label{sec:depth_vs_token_dimensions}

Input-independent queries give depth-wise routing a useful structural property: all query vectors lie in a fixed subspace. Let $Q\in\mathbb{R}^{d\times T}$ contain the query vectors at the $T$ active standard \AttnRes{} read sites, where $d$ is the residual width. Any coordinate-wise scaling from the routing RMSNorm is absorbed into these vectors. The rank of $Q$ counts the independent directions needed to represent all queries. Each routing score depends only on the projection of the normalized key onto the subspace spanned by these directions.

This observation gives an exact conversion from standard \AttnRes{} to \LRAttnRes{} with full-width RMS normalization under two assumptions. First, keys are normalized using all $d$ residual coordinates before selecting the $r$ routing coordinates. Second, the underlying Transformer allows its weights to be reparameterized consistently under an orthogonal change of residual coordinates. Such a change preserves vector lengths and hence full-width RMS normalization~\citep{ashkboos2024slicegpt}. We can therefore align the query subspace with $r$ retained coordinates whenever $r\geq\operatorname{rank}(Q)$, preserving every routing distribution and model output while keeping residual values at width $d$. Our 24-layer model has $T=48$, so there are at most 48 independent query directions and $r\geq48$ suffices regardless of source grouping. Appendix~\ref{app:static_query_span} states the precise assumptions and gives the proof.

We convert six \AttnRes checkpoints to \LRAttnRes without retraining (Table~\ref{tab:checkpoint_conversion}). The largest FP64 logit difference is $8.251\times10^{-13}$. Paired BF16 evaluations over 100M tokens yield cross-entropy (CE) increases ranging from $6.086\times10^{-5}$ to $8.865\times10^{-5}$. Appendix~\ref{app:query_span_verification} details verification. 

The basis change is absorbed into existing weights, adding no inference-time rotations or projections. An implementation exploiting the reduced query width lowers score dot products from $Sd$ to $Sr$ multiply-adds per read over $S$ sources. Thus conversion permits lower routing FLOPs without additional projection FLOPs.

This conversion could also be applied to large models such as Kimi K3: its 186 active folded queries have joint rank at most 186 despite value width $d=7168$ (Appendix~\ref{app:kimi_query_utilization}). A whole-model conversion would therefore permit scoring width $r=186$ with mathematically identical routing distributions and model outputs. We have not converted Kimi K3 or measured its inference speed due to computational and budget constraints.

\begin{table}[!ht]
    \centering
    \caption{Conversion without retraining. FP64 logit errors use one 2048-token sequence. Paired CE uses 100M tokens with BF16 model computation, FP32 per-token cross-entropy, and FP64 summation. $\Delta$CE is converted minus original before rounding.}
    \label{tab:checkpoint_conversion}
    \vspace{0.65em}
    \small
    \setlength{\tabcolsep}{4pt}
    \begin{tabular*}{\linewidth}{@{\extracolsep{\fill}}llrrrrr@{}}
        \toprule
        Model & Routing & $d\to r$ & \shortstack{FP64 max. error\\($\times10^{-13}$)} & \shortstack{Original\\CE} & \shortstack{Converted\\CE} & \shortstack{$\Delta$CE\\($\times10^{-5}$)} \\
        \midrule
        0.5B & Full & $1024\to64$ & $7.407$ & $2.972$ & $2.972$ & $+6.086$ \\
        0.5B & $N=4$ & $1024\to64$ & $5.853$ & $2.976$ & $2.976$ & $+8.865$ \\
        0.5B & $N=8$ & $1024\to64$ & $3.852$ & $2.974$ & $2.974$ & $+7.768$ \\
        0.5B & $N=16$ & $1024\to64$ & $3.300$ & $2.964$ & $2.964$ & $+7.908$ \\
        1B & $N=8$ & $1536\to64$ & $4.126$ & $2.800$ & $2.800$ & $+6.398$ \\
        4B & $N=8$ & $2560\to128$ & $8.251$ & $2.738$ & $2.738$ & $+6.998$ \\
        \bottomrule
    \end{tabular*}
\end{table}

\subsection{Training throughput}
\label{sec:training_throughput}

Table~\ref{tab:fast_attnres_throughput} reports complete training-step throughput over five paired repeats. Flash Linear Attention (FLA, \citealp{yang2024fla}) is an open-source GPU-kernel library. Its fused standard \AttnRes{} kernel is the fastest baseline we tested. Reducing rank from $d$ to $d/4$ with $N=8$ under \fastattnres{} gives mean paired gains of $0.59$--$1.20\%$. The table reports their $95\%$ confidence intervals. Combined kernel and rank gains over FLA are $1.55\%$ and $1.54\%$ at $1\text{B}$ and $4\text{B}$ on H100, and $2.87\%$ and $2.97\%$ on B200. Appendix~\ref{app:fast_attnres} gives the protocol, uncertainty calculation, and correctness checks.

\begin{table}[H]
    \centering
    \caption{Complete training-step throughput (tokens/s, mean $\pm$ sample SD, five repeats). Rank gains compare $r=d/4$ with $r=d$ under \fastattnres{} (paired mean and $95\%$ CI).}
    \label{tab:fast_attnres_throughput}
    \vspace{0.65em}
    \small
    \setlength{\tabcolsep}{4pt}
    \newcommand{\FastThroughputRows}{%
H100 & 1B & $47{,}436\pm 34$ & $47{,}887\pm 48$ & $48{,}171\pm 179$ & $0.594$ & $[0.030, 1.157]$ \\
H100 & 4B & $15{,}097\pm 14$ & $15{,}208\pm 17$ & $15{,}329\pm 17$ & $0.795$ & $[0.552, 1.038]$ \\\addlinespace[2pt]
B200 & 1B & $66{,}489\pm 17$ & $67{,}933\pm 358$ & $68{,}400\pm 309$ & $0.688$ & $[0.591, 0.786]$ \\
B200 & 4B & $22{,}856\pm 8$ & $23{,}256\pm 36$ & $23{,}536\pm 15$ & $1.203$ & $[1.060, 1.345]$ \\
}
    \begin{tabular*}{\linewidth}{@{\extracolsep{\fill}}llrrrrr@{}}
        \toprule
        GPU & Model & \shortstack{FLA\\$r=d$} & \shortstack{Fast\\$r=d$} & \shortstack{Fast\\$r=d/4$} & \shortstack{Rank gain\\($\%$)} & \shortstack{$95\%$ CI\\($\%$)} \\
        \midrule
        \FastThroughputRows
        \bottomrule
    \end{tabular*}
\end{table}

\section{Conclusion}
\label{sec:conclusion}

We introduced \LRAttnRes{}, which computes depth-routing scores from a low-dimensional coordinate slice while preserving full-width values, without auxiliary key projections. In the 32-setting $0.5\text{B}$ sweep, 27 of 28 low-rank configurations improve on their matched standard endpoint while reducing the counted routing FLOPs. The $N=8,r=128$ configuration improves loss across three seeds, and the gain persists in a full-width-RMS control. Our analysis establishes a sufficient width for preserving standard expressivity under full-width normalization, supported by numerical checks on six converted checkpoints, and characterizes the local routing feedback under both normalization choices. At $1\text{B}$ and $4\text{B}$, $r=d/4$ yields lower validation loss and higher average accuracy across eight benchmarks. We also provide \fastattnres{}, a fused implementation of standard and low-rank routing. These findings support selecting routing-key width independently of value width.

\subsection*{AI use statement}

In this work, we used generative AI tools for implementing methods, providing feedback on experimental methodology, developing theoretical arguments, and interpreting results. We have not used generative AI tools for originating research ideas, generating synthetic data sets, translation, or directly cleaning and reformatting data sets, and qualitative and thematic analyses are not applicable to this work. Additionally, we used generative AI tools for creating and editing software code, preparing scientific figures, drafting and editing the manuscript, and identifying relevant literature. All AI-assisted code, analyses, figures, and text underwent human review. We take responsibility for the final content of this work, including text, claims or artifacts produced with the aid of generative AI.

\subsection*{Reproducibility statement}

We provide the \LRAttnRes{} and \fastattnres{} codebases in the linked public repositories. Section~\ref{sec:method} and Appendix~\ref{app:lr_attnres_pseudocode} describe the routing mechanisms and provide implementation pseudocode, while Sections~\ref{sec:experimental_setup} and~\ref{sec:preliminary_1b} specify the model architectures and training setup. Appendix~\ref{app:static_query_span} states the assumptions underlying our theoretical results and provides proofs and numerical verification details. Appendix~\ref{app:costs} describes the FLOPs, memory, and parameter accounting, and Appendix~\ref{app:preliminary_1b} specifies the downstream evaluation protocol. Appendix~\ref{app:fast_attnres} documents the \fastattnres{} implementation, numerical correctness checks, and hardware throughput measurement procedure.

\bibliography{iclr2027_conference}
\bibliographystyle{iclr2027_conference}

\appendix

\section{Validation--FLOPs}
\label{app:validation_results}

\label{app:additional_validation}

Table~\ref{tab:additional_validation_results} presents validation losses and added routing FLOPs across full routing and block settings $N\in\{4,8,16\}$ at eight ranks $r\in\{16,32,64,128,256,512,768,1024\}$. At $r=d=1024$, each setting is standard \AttnRes{}. All runs use a 100M-token held-out validation set. Table~\ref{tab:validation_results_all} includes baseline and projected variants.

\begin{table}[!ht]
    \centering
    \caption{Validation losses and added routing FLOPs for all 32 configurations in Figure~\ref{fig:validation_vs_rank}. Added FLOPs are reported as a percentage of the non-embedding Transformer core. Bold marks the lowest loss.}
    \label{tab:additional_validation_results}
    \vspace{0.75em}
    \footnotesize
    \setlength{\tabcolsep}{4pt}
    \newcommand{\RoutingCostRows}{%
16 & 0.4129 & 2.973 & 0.05668 & 2.960 & 0.08906 & 2.958 & 0.1538 & 2.961 \\
32 & 0.4193 & 2.965 & 0.05755 & 2.953 & 0.09043 & 2.951 & 0.1562 & 2.949 \\
64 & 0.4320 & 2.960 & 0.05929 & 2.950 & 0.09317 & 2.945 & 0.1609 & 2.946 \\
128 & 0.4574 & 2.959 & 0.06278 & 2.948 & 0.09865 & \textbf{2.944} & 0.1704 & 2.947 \\
256 & 0.5082 & 2.959 & 0.06975 & 2.956 & 0.1096 & 2.956 & 0.1893 & 2.955 \\
512 & 0.6099 & 2.958 & 0.08371 & 2.970 & 0.1315 & 2.966 & 0.2272 & 2.953 \\
768 & 0.7115 & 2.962 & 0.09766 & 2.974 & 0.1535 & 2.969 & 0.2651 & 2.956 \\
1024 & 0.8131 & 2.972 & 0.1116 & 2.976 & 0.1754 & 2.974 & 0.3029 & 2.964 \\
}
    \begin{tabular}{@{}r*{4}{rr}@{}}
        \toprule
        & \multicolumn{2}{c}{Full} & \multicolumn{2}{c}{Block $N=4$} & \multicolumn{2}{c}{Block $N=8$} & \multicolumn{2}{c}{Block $N=16$} \\
        \cmidrule(lr){2-3}\cmidrule(lr){4-5}\cmidrule(lr){6-7}\cmidrule(lr){8-9}
        Rank $r$ & FLOPs (\%) & Val. loss & FLOPs (\%) & Val. loss & FLOPs (\%) & Val. loss & FLOPs (\%) & Val. loss \\
        \midrule
        \RoutingCostRows
        \bottomrule
    \end{tabular}
\end{table}

\paragraph{Seed repeats and normalization control.}
Across the three paired runs with different seeds, standard $N=8$ \AttnRes{} ($r=1024$) has validation losses $(2.974,\,2.972,\,2.973)$, and slice-normalized $N=8,r=128$ \LRAttnRes{} has losses $(2.944,\,2.946,\,2.943)$. All runs use the $0.5\text{B}$ architecture, $10\text{B}$ training tokens, and the same $100\text{M}$-token validation set. The full-width-RMS control reaches $2.943$. The sweep tables retain single-run sweep values.

\section{Static-query expressivity and normalization}
\label{app:static_query_span}

\subsection{A sufficient width from the joint query span}
\label{app:query_span_proof}

The following argument applies the orthogonal computational invariance of RMSNorm Transformers established by SliceGPT~\citep{ashkboos2024slicegpt}, also used in QuaRot~\citep{ashkboos2024quarot}, to static depth-routing queries. It concerns full-width RMS normalization, as in the control in Eq.~\ref{eq:rms_control}, while retaining all $d$ value coordinates.

\paragraph{Affine normalization.}
Routing RMSNorm gains can be folded into the static query: $\vq_t=\boldsymbol\gamma_t\odot\mathbf w_t$ gives
\begin{equation}
\mathbf w_t^\top\!\left(\boldsymbol\gamma_t\odot\frac{\vo}{\rho_d(\vo)}\right)
=\frac{\vq_t^\top\vo}{\rho_d(\vo)},
\qquad \rho_d(\vo)=\sqrt{\|\vo\|_2^2/d+\epsilon}.
\label{eq:query_span_affine_fold}
\end{equation}
Likewise, a backbone RMSNorm gain followed by a linear map $W$ can be absorbed into $\widehat W=W\operatorname{diag}(\boldsymbol\gamma)$, leaving scalar RMS normalization. This reduction requires that every consumer admit the absorption without violating parameter-sharing constraints. The proposition applies after these absorptions. $Q$ contains the effective routing queries.

\paragraph{Proposition 1 (Static-query span).}
Consider a deterministic \AttnRes{} decoder with scalar residual RMSNorm, unrestricted residual-facing linear projections, and $T$ input-independent routing queries. Assume that its embeddings, vocabulary head, and remaining residual-coordinate operations admit the same orthogonal reparameterization. Let $Q=[\vq_1,\ldots,\vq_T]\in\R^{d\times T}$ and $k=\operatorname{rank}(Q)$. For every $1\leq r\leq d$ with $r\geq k$, there is a consistent orthogonal reparameterization whose queries use only the final $r$ coordinates and whose routing distributions and vocabulary logits equal those of the original model for every input in exact arithmetic. The routing denominator remains full-width. The result holds for both full routing and the block sums defined in Section~\ref{sec:method}.

\paragraph{Proof.}
Let $P_r\in\R^{r\times d}$ select the final $r$ coordinates. Choose an orthogonal $R\in\R^{d\times d}$ that maps the column span of $Q$ into this coordinate subspace, and set $\vq'_t=P_rR\vq_t$ and $\vo'_i=R\vo_i$. With $\rho_d(\vo)=\sqrt{\|\vo\|_2^2/d+\epsilon}$, orthogonality gives
\begin{equation}
\rho_d(R\vo)=\rho_d(\vo),
\qquad
\rmsnorm(R\vo)=R\,\rmsnorm(\vo).
\label{eq:query_span_rms_invariance}
\end{equation}
Since $P_r^\top P_rR\vq_t=R\vq_t$, every routing score is preserved:
\begin{equation}
\frac{(\vq'_t)^\top P_r\vo'_i}{\rho_d(\vo'_i)}
=\frac{(R\vq_t)^\top P_r^\top P_rR\vo_i}{\rho_d(\vo_i)}
=\frac{\vq_t^\top\vo_i}{\rho_d(\vo_i)}.
\label{eq:query_span_score_equivalence}
\end{equation}
Thus the softmax weights are unchanged and the mixed value transforms as $\vh'_t=R\vh_t$. Block accumulation also commutes with $R$, since $\sum_{i\in B}R\vo_i=R\sum_{i\in B}\vo_i$.

For column-vector activations, transform each residual-facing input projection as $W'_{\mathrm{in}}=W_{\mathrm{in}}R^\top$ and each output projection as $W'_{\mathrm{out}}=RW_{\mathrm{out}}$. Rotate embeddings and any residual-output biases by $R$, and transform the vocabulary head as $W'_{\mathrm{head}}=W_{\mathrm{head}}R^\top$. Equation~\ref{eq:query_span_rms_invariance} then preserves the inputs to internal token-attention and SwiGLU computations, including QK normalization and RoPE. Their outputs rotate back into the new residual basis. Induction over residual reads therefore preserves every routing distribution and final vocabulary logit. All transformations are absorbed into existing parameters. $\square$

For a fixed checkpoint, its joint query rank $k$ suffices. Across the stated architecture, $k\leq\min(T,d)$ gives a uniform sufficient width. Conversely, a full-width query can reproduce any reduced query by zero-padding it, so the full-width-RMS function classes coincide for $r\geq\min(T,d)$. Our 24-layer decoder has $T=48$ active queries. The embedding-only read needs none. This bound counts queries, not candidate sources, and does not predict the validation-optimal rank. It also retains full-width normalization cost. Kimi K3's 186 active folded queries similarly have joint rank at most 186. Applying the whole-model construction additionally requires the stated backbone assumptions.

\paragraph{Minimal shared score projection.}
For fixed queries and arbitrary source vectors, $k$ is also necessary for an exact shared linear score projection retaining the full-width denominator. If $U\in\R^{d\times r}$ and reduced queries $\va_t$ satisfy $\va_t^\top U^\top\vo/\rho_d(\vo)=\vq_t^\top\vo/\rho_d(\vo)$ for every $\vo$, then $U\va_t=\vq_t$. Hence $\operatorname{col}(Q)\subseteq\operatorname{col}(U)$ and $r\geq k$. An orthonormal basis of $\operatorname{col}(Q)$ attains this bound. This characterizes exact score projection on arbitrary sources, not the minimum width of a retrained network with different representations.

\subsection{Numerical verification}
\label{app:query_span_verification}

The joint query spans have dimension 48 for the $0.5\text{B}$ and $1\text{B}$ checkpoints and 88 for $4\text{B}$. A shared orthogonal basis change places these spans within the final $r=64$ or $128$ coordinates, respectively, while retaining full-width values and the full-width RMS denominator. We verify FP64 logits on one complete 2048-token sequence per model, then export and strictly reload the BF16 checkpoints. Their native routing implementations match the reference on this sequence. Paired original and converted evaluations use the same 100M validation tokens, with FP32 per-token cross-entropy and FP64 summation. Table~\ref{tab:checkpoint_conversion} reports numerical fidelity without additional training.

\subsection{Local value and routing gradients}
\label{app:routing_feedback_proof}

Fix one read with $\vh=\sum_i\alpha_i\vo_i$, $\valpha=\softmax(\vz)$, and source-local scores $z_i=z_i(\vo_i)$ with the static query held fixed. All derivatives below are contributions through this read: source slots are treated as independent graph inputs. If a tensor also participates in other reads or in constructing later sources, its total gradient includes those additional paths.

\paragraph{Lemma 2 (Value and routing contributions).}
For upstream gradient $\vg=\nabla_{\vh}\mathcal L$, Eq.~\ref{eq:value_routing_gradient} holds. Indeed, the softmax Jacobian gives
\begin{equation}
\frac{\partial\vh}{\partial z_i}=\alpha_i(\vo_i-\vh),
\qquad
\mathrm d\vh=\alpha_i\,\mathrm d\vo_i+\alpha_i(\vo_i-\vh)\,\mathrm d z_i.
\label{eq:routing_feedback_differential}
\end{equation}
Taking the inner product with $\vg$ yields the decomposition. Equivalently, the local value Jacobian is $\partial\vh/\partial\vo_i=\alpha_i I+\alpha_i(\vo_i-\vh)(\nabla_{\vo_i}z_i)^\top$. $\square$

\paragraph{Corollary 2.1 (Slice-normalized routing).}
Let $\vk_i=P_r\vo_i$, $\rho_i=\rho_r(\vk_i)$, and $z_i^{\mathrm{slice}}=\vq^\top\vk_i/\rho_i$. Direct differentiation gives
\begin{equation}
\nabla_{\vo_i}z_i^{\mathrm{slice}}
=P_r^\top\left[\frac{\vq}{\rho_i}
-\frac{(\vq^\top\vk_i)\vk_i}{r\rho_i^3}\right].
\label{eq:slice_score_derivative}
\end{equation}
Its support lies in the selected coordinates. Thus $(I-\Pi_r)\nabla_{\vo_i}\mathcal L=\alpha_i(I-\Pi_r)\vg$. More strongly, any finite perturbation $\vu\in\ker P_r$ of source $i$ leaves all scores and weights at this read unchanged, while the output becomes $\vh+\alpha_i\vu$. This is local invariance to the unselected value coordinates. It does not prevent later layers from mixing them with selected coordinates.

\paragraph{Corollary 2.2 (Full-width-RMS routing).}
For $z_i^{\mathrm{control}}=\vq^\top P_r\vo_i/\rho_d(\vo_i)$,
\begin{equation}
\nabla_{\vo_i}z_i^{\mathrm{control}}
=\frac{P_r^\top\vq}{\rho_d(\vo_i)}
-\frac{z_i^{\mathrm{control}}\vo_i}{d\,\rho_d(\vo_i)^2}.
\label{eq:control_score_derivative}
\end{equation}
Projection onto the unselected coordinates gives Eq.~\ref{eq:routing_feedback_support}. Their contribution to the score is through their norm. The corresponding derivative is parallel to $(I-\Pi_r)\vo_i$. The control therefore preserves a radial dependence outside the slice.

\paragraph{Corollary 2.3 (Local value transmission).}
For finite logits, $\alpha_i>0$. By Lemma~2, the local source-to-output Jacobian satisfies
\begin{equation}
\operatorname{rank}\!\left(\frac{\partial\vh}{\partial\vo_i}\right)
=\operatorname{rank}\!\left[I_d+(\vo_i-\vh)(\nabla_{\vo_i}z_i)^\top\right]
\geq d-1,
\label{eq:local_value_jacobian_rank}
\end{equation}
since the bracket is the identity plus a matrix of rank at most one. Thus narrow scoring does not impose a rank-$r$ bottleneck on local value transmission under either normalization. This statement treats source slots independently at one read. It does not guarantee good conditioning or a full-rank end-to-end Jacobian.

These identities establish gradient support, not a reduction in feedback magnitude or harmful interference. The inner product between the direct and routing contributions is $\alpha_i^2\langle\vg,\vo_i-\vh\rangle\langle\vg,\nabla_{\vo_i}z_i\rangle$, which can have either sign. Standard full-width routing also has structured derivatives: $\nabla_{\vo_i}z_{t,i}$ lies in $\operatorname{span}(Q,\vo_i)$. Slicing fixes the coordinate support of the numerator's dependence, and slice normalization additionally removes the outside-slice norm dependence. Neither property alone determines validation loss.

\subsection{Block aggregation and routing geometry}
\label{app:block_routing_geometry}

At a fixed read, a completed block is exposed only through $\vb_B=\sum_{j\in B}\vo_j$. Perturbations whose sum within that block is zero leave its value and routing score unchanged. Every constituent write has the same explicit mixture coefficient $\alpha_B$ and receives the same adjoint through the block-sum node. These statements hold for that read with the other source slots fixed. They do not identify the complete gradients of the original writes across the network.

The normalization of a block depends on cross terms between its writes:
\begin{equation}
\left\|P_r\vb_B\right\|_2^2
=\sum_{j,k\in B}(P_r\vo_j)^\top(P_r\vo_k),
\qquad
\left\|\vb_B\right\|_2^2
=\sum_{j,k\in B}\vo_j^\top\vo_k.
\label{eq:block_norm_cross_terms}
\end{equation}
Alignment and cancellation can differ between the selected coordinates and the full value space. Grouping therefore changes both the distinctions available to the router and the selected-to-full RMS ratio. The number of active queries, and hence the uniform sufficient-width bound in Proposition~1, is unchanged by grouping. Fewer candidate sources alone do not imply a smaller necessary query span or predict the best rank of a model trained from scratch.

\FloatBarrier
\section{Projected Low-Rank Attention Residuals}
\label{app:projected_lr}

\PLRAttnRes explores exogenous routing keys generated by auxiliary output projections. Here, exogenous means separate from the residual-value coordinates. The keys still depend on the same sub-layer activations. This provides an interesting test of dedicated routing capacity, but its additional projections, parameters, and key storage make it a less practical choice for LLM training than \LRAttnRes.

\subsection{Projected low-rank keys}
\label{sec:projected_lr_keys}

\PLRAttnRes uses a learned low-rank key emitted from the same output projection that produces the residual value. Let $\va_i\in\R^{m_i}$ be the activation passed to the output projection of residual-writing sub-layer $i$. For an attention sub-layer, this is the activation passed to the attention output projection, so typically $m_i=d$. For a SwiGLU feed-forward sub-layer, this is the activation passed to the second/output projection of the FFN, so $m_i$ is the SwiGLU hidden width. If the ordinary output projection is $W_i^O\in\R^{d\times m_i}$, the standard residual contribution is
\begin{equation}
    \vo_i = W_i^O \va_i .
    \label{eq:standard_output_projection}
\end{equation}
\PLRAttnRes adds a low-rank key projection and computes the output and key together:
\begin{equation}
    \begin{bmatrix}
        \vo_i \\
        \vk_i^{\mathrm{P}}
    \end{bmatrix}
    =
    \begin{bmatrix}
        W_i^O \\
        W_i^K
    \end{bmatrix}
    \va_i,
    \qquad
    W_i^O\in\R^{d\times m_i},
    \qquad
    W_i^K\in\R^{r\times m_i},
    \qquad
    \vo_i\in\R^d,
    \qquad
    \vk_i^{\mathrm{P}}\in\R^r.
    \label{eq:fused_projection}
\end{equation}
In the attention sub-layer, the fused matrix augments the attention output projection. In the SwiGLU feed-forward sub-layer, it augments the second/output projection that maps the gated hidden activation back to the model dimension.

The embedding source has no sub-layer output projection. The default \PLRAttnRes embedding key is therefore produced by a small low-rank projection:
\begin{equation}
    \vo_0 = \operatorname{Embed}(x),
    \qquad
    \vk_0^{\mathrm{P}} = W_0^K \vo_0,
    \qquad
    \bar{\vk}_0^{\mathrm{P}} = \rmsnorm(\vk_0^{\mathrm{P}}),
    \qquad
    W_0^K\in\R^{r\times d}.
    \label{eq:embedding_key_projection}
\end{equation}

\PLRAttnRes is an exploratory alternative to the primary method. Because the key has its own learned projection, the model can use the full value vector $\vo_i$ for residual-stream content while using a separate representation for routing. Its best block configuration achieves a validation loss of $2.944$. Its drawback is that the auxiliary key projections add projection FLOPs and parameters. The detailed accounting is given in Appendix~\ref{app:costs}.

\subsection{Inference compatibility and practical costs}

\PLRAttnRes preserves the same input-independent queries and online-softmax merge as \LRAttnRes, so it also satisfies the mathematical conditions for the predecessor's two-phase inference strategy. Its practical costs differ: the auxiliary projections add parameters and computation, and the retained source representation includes separate keys. Under a pipeline layout that forwards each key alongside its value, the payload per source is $d+r$ rather than $d$ elements, an increase of $3.125\%$ for $d=1024,r=32$. Appendix~\ref{app:costs} gives the full cost accounting.

\subsection{Complete validation--FLOPs comparison}
\label{app:projected_results}

\begin{table}[!htp]
    \centering
    \caption{
    Validation loss and routing cost for the baseline, standard \AttnRes{}, and all low-rank configurations. Bold marks the lowest loss within each method. Added FLOPs are reported as a percentage of the non-embedding Transformer core. For \PLRAttnRes{}, this column includes the auxiliary learned key-projection path. For \AttnRes{} and \LRAttnRes{}, it is the depth-wise residual kernel cost because no projected-key path is present. The default validation split contains $100\mathrm{M}$ held-out tokens.
    }
    \vspace{0.75em}
    \label{tab:validation_results_all}
    \small
    \setlength{\tabcolsep}{4pt}
    \newcommand{\ValidationComparisonRows}{%
Baseline Transformer & -- & -- & 0 & 2.997 \\
\midrule
Full \AttnRes{} & Full & 1024 & 0.8131 & 2.972 \\
Block \AttnRes{} & 4 & 1024 & 0.1116 & 2.976 \\
Block \AttnRes{} & 8 & 1024 & 0.1754 & 2.974 \\
Block \AttnRes{} & 16 & 1024 & 0.3029 & \textbf{2.964} \\
\midrule
Full \PLRAttnRes{} & Full & 16 & 0.8966 & 2.957 \\
Full \PLRAttnRes{} & Full & 32 & 1.387 & 2.950 \\
Full \PLRAttnRes{} & Full & 64 & 2.367 & 2.947 \\
Block \PLRAttnRes{} & 4 & 32 & 1.025 & 2.945 \\
Block \PLRAttnRes{} & 8 & 32 & 1.058 & \textbf{2.944} \\
Block \PLRAttnRes{} & 16 & 32 & 1.123 & 2.951 \\
\midrule
Full \LRAttnRes{} & Full & 16 & 0.4129 & 2.973 \\
Full \LRAttnRes{} & Full & 32 & 0.4193 & 2.965 \\
Full \LRAttnRes{} & Full & 64 & 0.4320 & 2.960 \\
Full \LRAttnRes{} & Full & 128 & 0.4574 & 2.959 \\
Full \LRAttnRes{} & Full & 256 & 0.5082 & 2.959 \\
Full \LRAttnRes{} & Full & 512 & 0.6099 & 2.958 \\
Full \LRAttnRes{} & Full & 768 & 0.7115 & 2.962 \\
Block \LRAttnRes{} & 4 & 16 & 0.05668 & 2.960 \\
Block \LRAttnRes{} & 4 & 32 & 0.05755 & 2.953 \\
Block \LRAttnRes{} & 4 & 64 & 0.05929 & 2.950 \\
Block \LRAttnRes{} & 4 & 128 & 0.06278 & 2.948 \\
Block \LRAttnRes{} & 4 & 256 & 0.06975 & 2.956 \\
Block \LRAttnRes{} & 4 & 512 & 0.08371 & 2.970 \\
Block \LRAttnRes{} & 4 & 768 & 0.09766 & 2.974 \\
Block \LRAttnRes{} & 8 & 16 & 0.08906 & 2.958 \\
Block \LRAttnRes{} & 8 & 32 & 0.09043 & 2.951 \\
Block \LRAttnRes{} & 8 & 64 & 0.09317 & 2.945 \\
Block \LRAttnRes{} & 8 & 128 & 0.09865 & \textbf{2.944} \\
Block \LRAttnRes{} & 8 & 256 & 0.1096 & 2.956 \\
Block \LRAttnRes{} & 8 & 512 & 0.1315 & 2.966 \\
Block \LRAttnRes{} & 8 & 768 & 0.1535 & 2.969 \\
Block \LRAttnRes{} & 16 & 16 & 0.1538 & 2.961 \\
Block \LRAttnRes{} & 16 & 32 & 0.1562 & 2.949 \\
Block \LRAttnRes{} & 16 & 64 & 0.1609 & 2.946 \\
Block \LRAttnRes{} & 16 & 128 & 0.1704 & 2.947 \\
Block \LRAttnRes{} & 16 & 256 & 0.1893 & 2.955 \\
Block \LRAttnRes{} & 16 & 512 & 0.2272 & 2.953 \\
Block \LRAttnRes{} & 16 & 768 & 0.2651 & 2.956 \\
\midrule
Block \LRAttnRes{} (full RMS) & 8 & 128 & 0.09865 & \textbf{2.943} \\
}
    \begin{tabular}{lrrrr}
        \toprule
        Method & Block count $N$ & Rank $r$ & Added FLOPs (\%) & Val. loss \\
        \midrule
        \ValidationComparisonRows
        \bottomrule
    \end{tabular}
\end{table}

The best \PLRAttnRes{} configuration uses block routing with $N=8$ and $r=32$, reaching $2.944$ validation loss at $1.058\%$ added FLOPs. At the same rank, projected block losses are $2.945$ for $N=4$ and $2.951$ for $N=16$. Full projected routing improves from $2.957$ to $2.947$ as rank increases from $16$ to $64$, while added FLOPs rise from $0.8966\%$ to $2.367\%$. Thus the projected results also depend on both rank and source grouping.

\begin{figure}[H]
    \centering
    \includegraphics[width=0.90\linewidth]{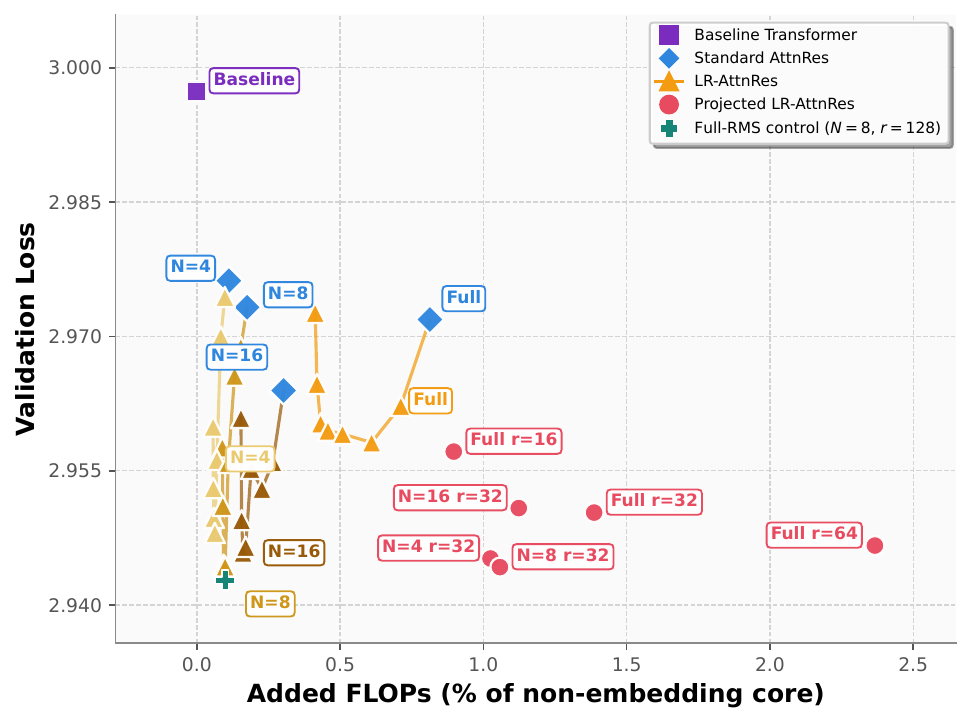}
    \caption{
    Validation loss versus added FLOPs relative to the non-embedding Transformer core at $0.5\text{B}$, including the full-width-RMS control (teal plus). Only the $N=8,r=128$ \LRAttnRes{} triangle and $N=8,r=d=1024$ standard \AttnRes{} diamond plot means across three training seeds, as in Figure~\ref{fig:validation_vs_rank}; all other points show single runs. Projected variants include auxiliary key-projection FLOPs. Other variants use the depth-wise score-and-value accounting in Appendix~\ref{app:costs}. Tables~\ref{tab:validation_results_all} and~\ref{tab:additional_validation_results} retain single-run sweep values.
    }
    \label{fig:validation_vs_depthwise_flops_all}
\end{figure}

The sliced sweep identifies $N=8,r=128$ as its lowest-loss configuration: $2.944$ validation loss at $0.09865\%$ added FLOPs. Its loss is $6.263\times10^{-4}$ lower than the best projected result, with $10.72\times$ lower added residual-side FLOPs. The sliced $N=8,r=64$ alternative has loss only $8.978\times10^{-4}$ higher than the sliced minimum, at $0.09317\%$ added FLOPs. It remains a lower-compute near-tie. At matched full-routing rank $r=64$, however, \PLRAttnRes{} reaches $2.947$ compared with $2.960$ for \LRAttnRes{}, a loss difference of $0.01365$. Dedicated keys therefore have lower observed loss in this full-routing comparison, while the block sweep offers a similar best loss at much lower routing cost (Table~\ref{tab:validation_results_all}).

\subsection{Projected routing diagnostics}
\label{app:projected_routing}

\begin{figure}[t]
    \centering
    \includegraphics[width=0.48\linewidth]{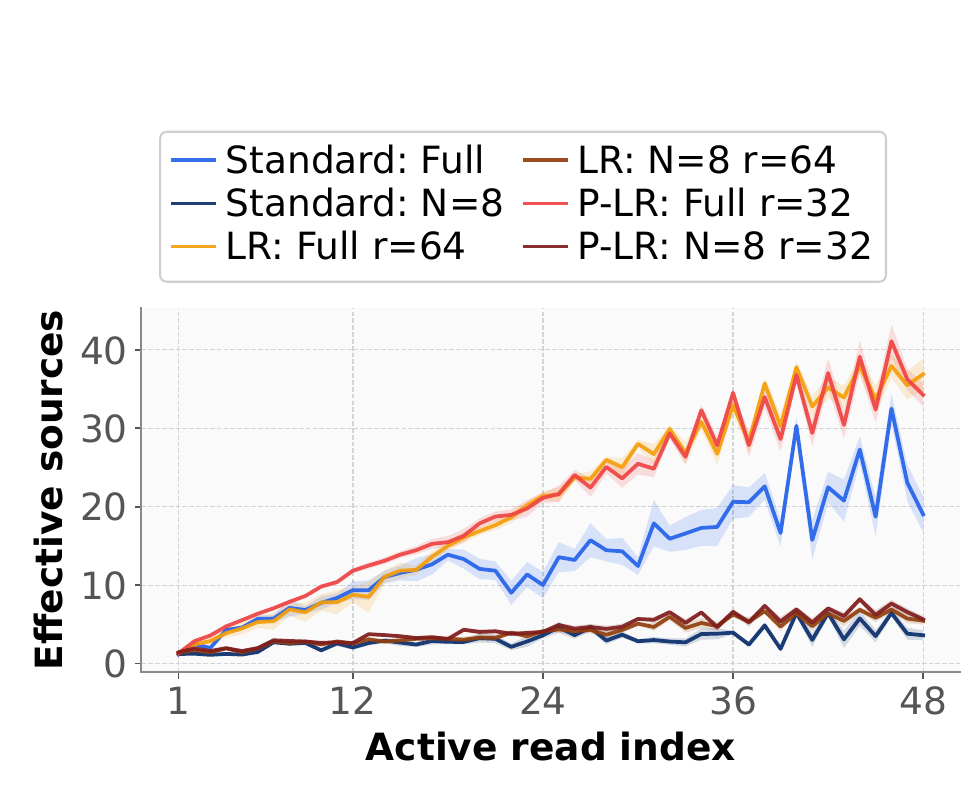}
    \hfill
    \includegraphics[width=0.48\linewidth]{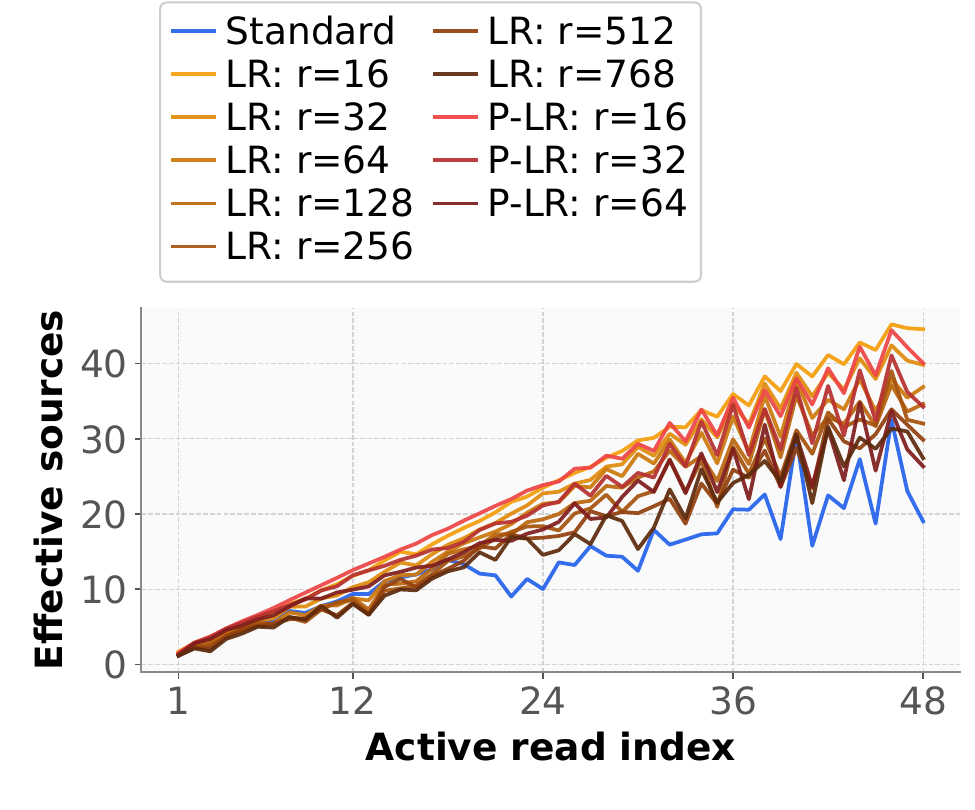}
    \caption{
    Effective number of residual sources across residual read sites. Left: comparison between standard \AttnRes{} (Standard), \LRAttnRes{} (LR), and \PLRAttnRes{} (P-LR) under full and block routing. Right: rank sweep for full \LRAttnRes{} and full \PLRAttnRes{}, with standard full \AttnRes{} shown in blue. Lines show token means; shading in the left panel indicates $\pm1$ standard deviation across $1\mathrm{M}$ held-out tokens. Higher values indicate broader aggregation.
    }
    \label{fig:effective_sources_all}
\end{figure}

Figure~\ref{fig:effective_sources_all} compares effective source counts across standard, projected, and sliced routing configurations, using the definition in Eq.~\ref{eq:effective_sources}. These routing diagnostics use $1{,}007{,}616$ held-out tokens. The validation losses in Table~\ref{tab:validation_results_all} come from a separate $100\mathrm{M}$-token cross-entropy evaluation with the same checkpoint settings.

The matched comparisons show broader source use with low-rank keys, but broader routing does not consistently predict lower validation loss (Section~\ref{sec:effective_sources}). Our interpretation is that restricting the routing representation may limit sensitivity to incidental value features while preserving useful source distinctions. 

\subsection{Gradient norms across routing configurations}
\label{app:projected_gradients}
\label{sec:grad_norm_stability}

\begin{figure}[!ht]
    \centering
    \includegraphics[width=0.90\linewidth]{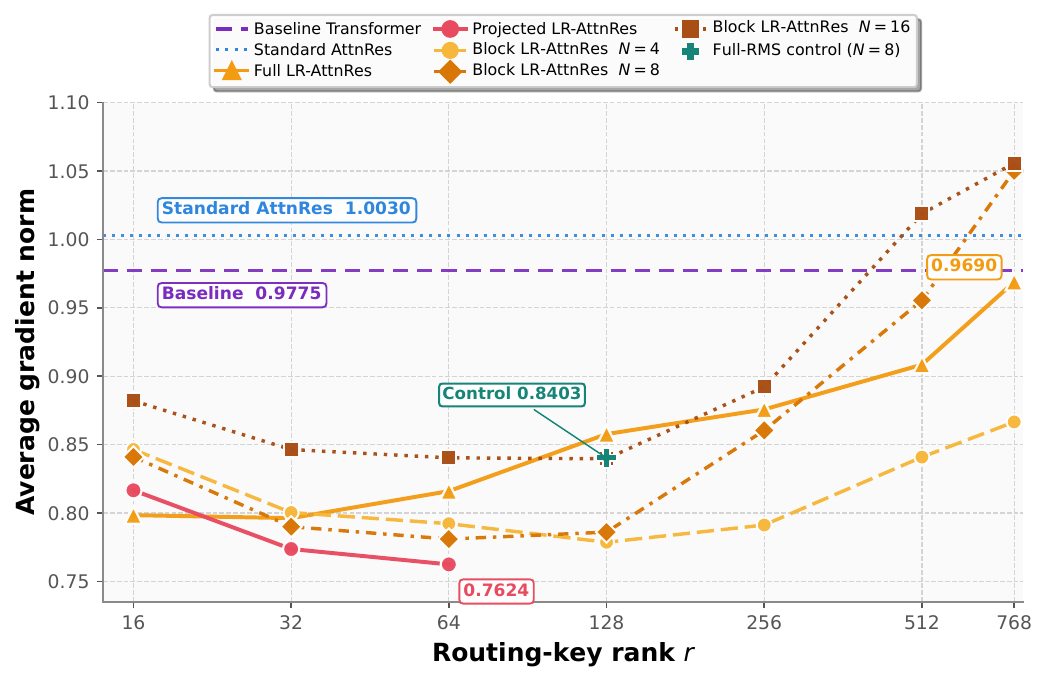}
    \caption{
    Average gradient norm by routing-key rank for the baseline, full \AttnRes{}, full low-rank variants, and block \LRAttnRes{}. The teal plus marks the $N=8,r=128$ full-width-RMS control. The purple dashed line shows the baseline. The blue dotted line shows full standard \AttnRes{}. Curves connect ranks within the same low-rank construction and block count $N$. 
    }
    \label{fig:grad_norm_stability_all}
\end{figure}

Figure~\ref{fig:grad_norm_stability_all} compares mean gradient norms across routing-key ranks for projected and sliced routing, including the full-width-RMS control.

For full \PLRAttnRes{}, the mean gradient norm decreases from $0.8166$ to $0.7737$ and $0.7624$ at ranks $16$, $32$, and $64$, respectively, compared with $1.003$ for full standard \AttnRes{}. These three means are also below the baseline mean of $0.9775$. The sliced sweep has a different response: full routing reaches its smallest mean at $r=32$, whereas block settings $N=4,8,16$ reach theirs at $r=128,64,128$. The sliced $N=8$ validation minimum is instead at $r=128$. Gradient magnitude and final validation loss therefore need not select the same rank.

For full sliced routing, the mean norm rises from $0.7963$ at $r=32$ to $0.9690$ at $r=768$. At $r=768$, the $N=8$ and $N=16$ means reach $1.050$ and $1.056$, exceeding full routing. The nearly equal block losses at $r=256$ accompany distinct mean norms of $0.7912$, $0.8604$, and $0.8921$ for $N=4,8,16$. Similar final losses therefore need not imply identical gradient statistics.

The full-routing validation minimum occurs at $r=512$, compared with $r=32$ for its mean gradient norm. The full-width-RMS control has a higher mean norm than sliced $N=8,r=128$ ($0.8403$ versus $0.7861$), despite its lower loss. Lower gradient norms accompany some validation gains, but do not necessarily explain them.

Our interpretation is that a separate projected key may ease the demands placed on the routing representation, while reusing value coordinates may introduce competing routing and representation requirements. These are hypotheses about routing difficulty and representational tension, not quantities measured by gradient norms. In particular, the non-monotonic sliced curves do not support a general claim that increasing $r$ necessarily increases this tension.

\FloatBarrier

\section{Additional Routing Analyses}
\label{app:additional_routing}

\subsection{Mean attention heatmaps}
\label{sec:attention_heatmaps}

\begin{figure}[!t]
    \centering
    \includegraphics[width=\linewidth]{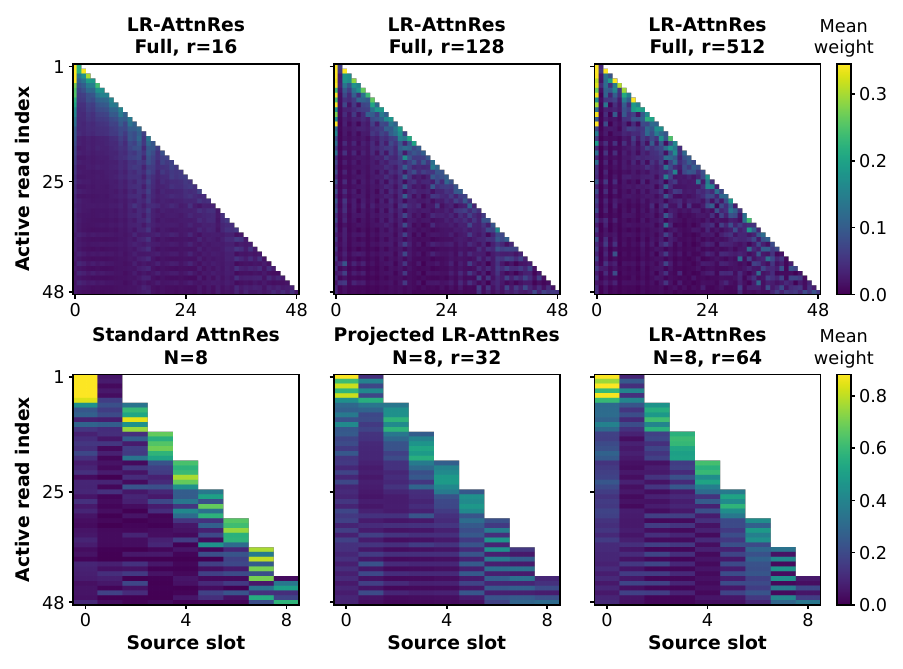}
    \caption{
    Mean depth-wise attention heatmaps over $1\mathrm{M}$ held-out tokens. Top row: full \LRAttnRes{} with $r=16$, $r=128$, and $r=512$. Bottom row: standard \AttnRes{}, \PLRAttnRes{} ($r=32$), and \LRAttnRes{} ($r=64$), all with $N=8$. The rows use separate color scales, with fixed upper limits of $0.345$ and $0.883$, respectively.
    }
    \label{fig:attention_mean_heatmaps}
\end{figure}

To complement the entropy-based effective-source analysis, we also visualize the mean depth-wise attention weights directly. For each residual read site, we average the source-attention distribution over validation tokens and plot the resulting source-by-depth pattern. Figure~\ref{fig:attention_mean_heatmaps} shows full \LRAttnRes{} at several ranks in the first row and block variants with $N=8$ in the second row.

The full \LRAttnRes{} heatmaps show a clear rank-dependent smoothness pattern. At very small rank, especially $r=16$, the attention map is comparatively smooth: probability mass changes gradually across residual read sites, and the model tends to form broad depth preferences rather than highly source-specific peaks. 

As the rank increases to $r=128$ and $r=512$, the heatmaps become visibly less smooth. Checkerboard-like patterns, localized spikes, and sharper structures appear. This accompanies the reduction in effective-source count for the full sliced rank sweep (Figure~\ref{fig:effective_sources}). Validation loss improves from $r=16$ to $r=512$ despite the less smooth mean patterns.

All three block heatmaps show stage-dependent source preferences. At the final read, standard $N=8$ \AttnRes{}, projected $N=8,r=32$, and sliced $N=8,r=64$ routing have token-averaged effective-source counts of $3.62$, $5.65$, and $5.47$, respectively. The heatmaps average attention weights across tokens; they do not establish smoother routing for individual tokens or a causal explanation for validation-loss differences.

\section{Kimi K3 folded-query utilization}
\label{app:kimi_query_utilization}

\subsection{Query construction and selection}

Kimi K3 has hidden width $d=7168$ and 93 Transformer layers partitioned into $N=8$ \AttnRes{} blocks~\citep{kimik3}. At each residual-routing site, the implementation compares the current within-block prefix with the retained summaries of completed blocks and the input embedding. The deepest routing operations therefore select among at most nine source vectors.

The checkpoint stores one pre-attention query and one pre-MLP query for each layer, plus one final-output query, giving 187 stored query sites. The first layer's pre-attention query is structurally unused because only one source (input embedding) exists and the official forward pass skips that routing call. We therefore exclude it and analyze 186 active queries: 92 pre-attention queries, 93 pre-MLP queries, and one final-output query.

The implementation folds the RMSNorm affine weight into the learned query projection before computing routing scores. If $\boldsymbol{\gamma}_t\in\R^d$ is the RMSNorm weight and $\mathbf{w}_t\in\R^d$ is the projection row at site $t$, the operational score vector is
\begin{equation}
    \widetilde{\mathbf{q}}_t
    =
    \boldsymbol{\gamma}_t \odot \mathbf{w}_t.
    \label{eq:kimi_folded_query}
\end{equation}
Both factors are explicitly cast to FP32 before the elementwise product in the released implementation. All results below use this folded FP32 vector.

\subsection{Coordinate effective dimension and exact zeros}

For each folded query, we convert squared coordinate magnitudes into a probability distribution,
\begin{equation}
    p_{t,j}
    =
    \frac{\widetilde{q}_{t,j}^{\,2}}
    {\sum_{k=1}^{d}\widetilde{q}_{t,k}^{\,2}},
\end{equation}
and define its Shannon coordinate effective dimension as
\begin{equation}
    D_{\mathrm{eff}}(\widetilde{\mathbf{q}}_t)
    =
    \exp\!\left(
        -\sum_{j=1}^{d}p_{t,j}\log p_{t,j}
    \right).
    \label{eq:kimi_coordinate_dimension}
\end{equation}
This is the entropy perplexity of the coordinate-energy distribution: it equals $d$ when energy is uniform over all coordinates and equals $1$ when one coordinate contains all energy.

\begin{figure}[t]
    \centering
    \includegraphics[width=\linewidth]{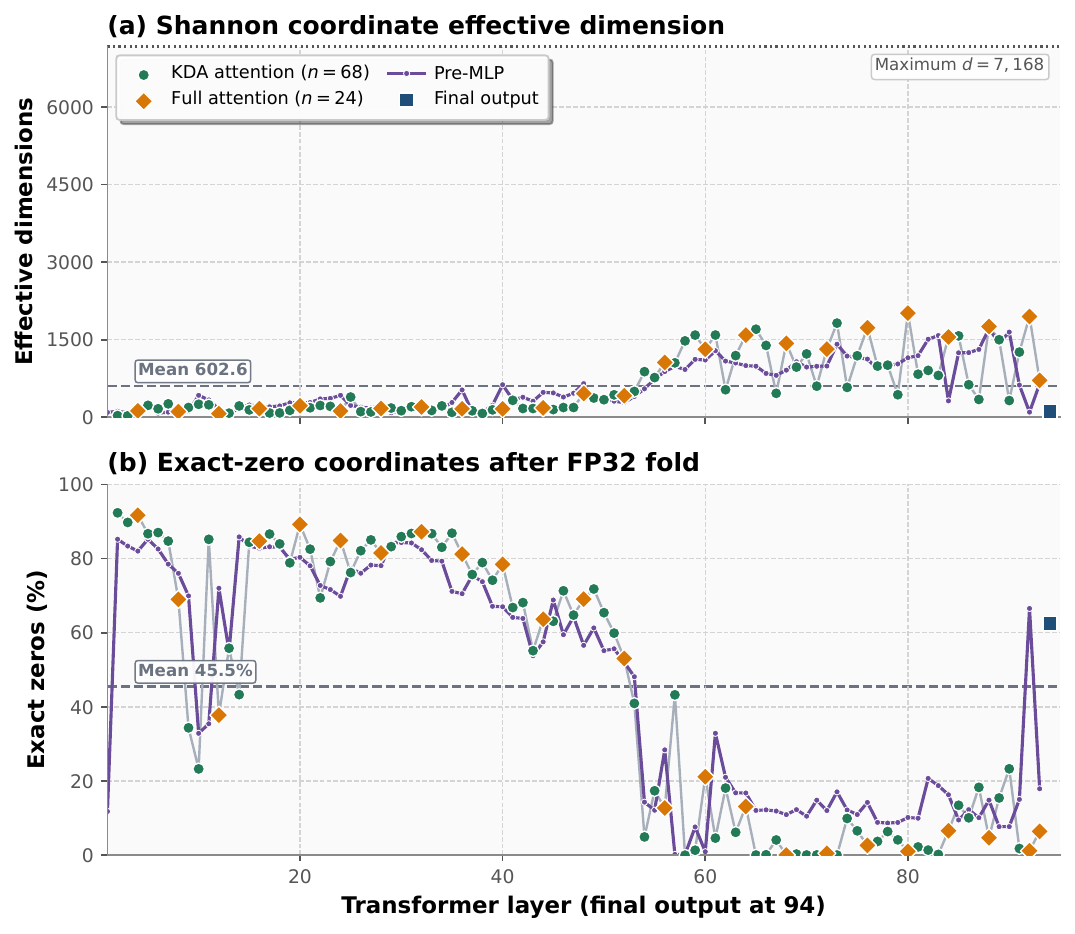}
    \caption{
    Layerwise utilization of Kimi K3's 186 active folded \AttnRes{} queries. Top: Shannon coordinate effective dimension from Equation~\ref{eq:kimi_coordinate_dimension}, shown on the full $0$--$7168$ width scale. Bottom: fraction of coordinates exactly zero after the FP32 fold. Dashed lines show the all-query means of $602.6$ effective dimensions and $45.5\%$ exact zeros.
    }
    \label{fig:kimi_query_utilization}
\end{figure}

\begin{table}[!ht]
    \centering
    \caption{
    Folded-query utilization by site type. KDA denotes Kimi Delta Attention. Full denotes full token attention. Effective-dimension statistics are computed per query. Zero fractions are averaged over queries in each row.
    }
    \vspace{0.5em}
    \label{tab:kimi_query_utilization}
    \begin{tabular}{lrrrr}
        \toprule
        Query group & Count & Mean $D_{\mathrm{eff}}$ & Median $D_{\mathrm{eff}}$ & Mean zeros (\%) \\
        \midrule
        KDA pre-attention & 68 & 546.2 & 325.7 & 45.3 \\
        Full pre-attention & 24 & 791.5 & 437.3 & 43.4 \\
        Pre-MLP & 93 & 600.4 & 402.3 & 46.0 \\
        Final output & 1 & 111.8 & 111.8 & 62.5 \\
        \midrule
        All active queries & 186 & 602.6 & 369.7 & 45.5 \\
        \bottomrule
    \end{tabular}
\end{table}

We separately measure the fraction of each folded query's $d=7168$ coordinates that are exactly equal to zero, using no tolerance threshold. The layerwise results show that the folded static queries use substantially less than their full coordinate width on average, but $D_{\mathrm{eff}}$ should not be interpreted as a recommended low-rank width.

\subsection{Numerical precision and limitations}

Across the $186\times7168=1{,}333{,}248$ folded coordinates, 606{,}351 are exactly zero in the reconstructed FP32 product, giving the reported $45.4792\%$. Of those zeros, 606{,}264 occur even though both BF16-decoded factors are nonzero (they arise from underflow in the FP32 multiplication). Recomputing the same products in FP64 leaves only 87 exact zeros, confirming that almost all observed zeros arise during the FP32 fold.

Many learned folded-query coordinates are therefore exactly zero under the released implementation's arithmetic and make no direct contribution to the routing-score numerator. This is consistent with optimization suppressing some coordinates' contribution to source selection, suggesting that full-width routing queries may be unnecessary. Realizing memory or speed benefits requires an implementation that exploits this sparsity.

\section{FLOPs, activation memory, and parameter accounting}
\label{app:costs}

This appendix gives the detailed accounting for standard \AttnRes, projected low-rank keys, and sliced low-rank keys. The main distinction is between two types of cost. The \emph{depth-wise residual kernel} is the attention-over-depth computation itself: it scores retained sources and mixes their values. The \emph{projected-key path} is the extra work needed only by \PLRAttnRes to produce learned low-rank keys.

\begin{table}[!ht]
    \centering
    \caption{
    Residual-side cost summary. $S_{\mathrm{src}}=\sum_{t\in\mathcal{T}}|\mathcal{S}_t|$ is the total number of source reads across residual sites, and $K_{\mathrm{src}}$ is the number of retained source activations.
    }
    \vspace{0.75em}
    \label{tab:appendix_cost_summary}
    \small
    \setlength{\tabcolsep}{4pt}
    \begin{tabular}{@{}lccc@{}}
        \toprule
        Method & Total residual-side FLOPs & Source-cache activations & Added parameters \\
        \midrule
        \AttnRes{}
        &
        $2dS_{\mathrm{src}}$
        &
        $K_{\mathrm{src}}d$
        &
        $2Ld$
        \\
        \shortstack[l]{\textsc{Projected}\\\LRAttnRes{}}
        &
        $(d+r)S_{\mathrm{src}} + r[L(d+m)+d]$
        &
        $K_{\mathrm{src}}(d+r)$
        &
        $r[L(d+m)+d+2L]$
        \\
        \LRAttnRes{}
        &
        $(d+r)S_{\mathrm{src}}$
        &
        $K_{\mathrm{src}}d$
        &
        $2Lr$
        \\
        \bottomrule
    \end{tabular}
\end{table}

Let $\mathcal{T}$ be the set of residual read sites and let
\begin{equation}
    S_{\mathrm{src}}
    =
    \sum_{t\in\mathcal{T}} n_t,
    \qquad
    n_t = |\mathcal{S}_t|.
\end{equation}
The source-count term $S_{\mathrm{src}}$ depends on whether the method uses full or block sources.

\subsection{Depth-wise residual kernel FLOPs}

Ordinary \AttnRes uses the same $d$-dimensional tensor as both key and value. At residual site $t$, the score computation costs $n_t d$ multiply-adds and the weighted value mixture costs another $n_t d$ multiply-adds:
\begin{equation}
    C_{\mathrm{AttnRes}}(t)
    =
    n_t d + n_t d
    =
    2n_t d.
\end{equation}
Summing over residual read sites gives
\begin{equation}
    C_{\mathrm{AttnRes}}
    =
    2dS_{\mathrm{src}}.
    \label{eq:app_attnres_kernel_flops}
\end{equation}

Both \PLRAttnRes and \LRAttnRes use an $r$-dimensional key and a full $d$-dimensional value. Therefore the depth-wise residual kernel cost is
\begin{equation}
    C_{\mathrm{LR,kernel}}(t)
    =
    n_t r + n_t d
    =
    n_t(d+r),
\end{equation}
and
\begin{equation}
    C_{\mathrm{LR,kernel}}
    =
    (d+r)S_{\mathrm{src}}.
    \label{eq:app_lr_kernel_flops}
\end{equation}
The percentage reduction in the depth-wise residual kernel relative to ordinary \AttnRes is
\begin{align}
    \%\,\Delta C_{\mathrm{kernel}}
    &=
    100\left(
    1-\frac{C_{\mathrm{LR,kernel}}}{C_{\mathrm{AttnRes}}}
    \right) \\
    &=
    100\left(
    1-\frac{(d+r)S_{\mathrm{src}}}{2dS_{\mathrm{src}}}
    \right) \\
    &=
    100\left(\frac{d-r}{2d}\right).
    \label{eq:app_lr_kernel_decrease}
\end{align}
The full-width-RMS control retains the same $r$-dimensional dot products and $d$-dimensional value mixtures, so it has the same $(d+r)S_{\mathrm{src}}$ cost as slice-normalized \LRAttnRes{} under this accounting. Its RMS statistic instead uses all $d$ coordinates. This normalization work is outside the score-and-value count plotted in Figure~\ref{fig:validation_vs_depthwise_flops_all}.

The source-count term cancels, so the same kernel percentage applies to full and block source sets. These counts include score dot products and value mixing and exclude normalization and softmax arithmetic. Slice normalization additionally reduces the normalization width from $d$ to $r$, whereas full-width-RMS conversion retains the original normalization width.

\subsection{Projected-key FLOPs}

\PLRAttnRes produces a learned low-rank key by adding $r$ rows to output projections that already exist. Let $m_i$ be the input width of the output projection for residual-writing sub-layer $i$. The extra projected-key cost is
\begin{equation}
    C_{\mathrm{P,key\text{-}proj}}
    =
    r\sum_{i=1}^{2L}m_i
    +
    rd,
    \label{eq:app_projected_key_projection_flops_general}
\end{equation}
where the final $rd$ term is the embedding key projection. For a Transformer block with one attention output projection of input width $d$ and one SwiGLU output projection of input width $m$,
\begin{equation}
    \sum_{i=1}^{2L}m_i
    =
    L(d+m),
\end{equation}
so
\begin{equation}
    C_{\mathrm{P,key\text{-}proj}}
    =
    r\bigl[L(d+m)+d\bigr].
    \label{eq:app_projected_key_projection_flops}
\end{equation}

Thus the total residual-side cost of \PLRAttnRes is
\begin{equation}
    C_{\mathrm{P,total}}
    =
    (d+r)S_{\mathrm{src}}
    +
    r\bigl[L(d+m)+d\bigr].
    \label{eq:app_projected_total_flops}
\end{equation}
Relative to ordinary \AttnRes, the net percentage change is
\begin{equation}
    \%\,\Delta C_{\mathrm{P,total}}
    =
    100\left(
    1 -
    \frac{
    (d+r)S_{\mathrm{src}}
    +
    r\bigl[L(d+m)+d\bigr]
    }{
    2dS_{\mathrm{src}}
    }
    \right).
    \label{eq:app_projected_net_change}
\end{equation}
Therefore, \PLRAttnRes reduces the depth-wise residual kernel cost, but it reduces \emph{total} residual-side FLOPs only when
\begin{equation}
    r\bigl[L(d+m)+d\bigr]
    <
    (d-r)S_{\mathrm{src}}.
    \label{eq:app_projected_reduction_condition}
\end{equation}
This condition can fail, especially in block mode where $S_{\mathrm{src}}$ is small. In that case, \PLRAttnRes improves validation loss but does not reduce total added FLOPs once projected-key computation is included.

\subsection{Sliced-key FLOPs}

\LRAttnRes obtains its key by slicing the last $r$ dimensions of the already-computed residual value:
\begin{equation}
    \vk_i^{\mathrm{S}}
    =
    \tailr(\vo_i).
\end{equation}
It therefore has no auxiliary key projection:
\begin{equation}
    C_{\mathrm{S,key\text{-}proj}}
    =
    0.
\end{equation}
Its total residual-side cost is just the low-rank depth-wise residual kernel:
\begin{equation}
    C_{\mathrm{S,total}}
    =
    (d+r)S_{\mathrm{src}}.
    \label{eq:app_sliced_total_flops}
\end{equation}
Thus \LRAttnRes reduces total residual-side FLOPs relative to ordinary \AttnRes whenever $r<d$:
\begin{equation}
    \%\,\Delta C_{\mathrm{S,total}}
    =
    100\left(
    1 -
    \frac{(d+r)S_{\mathrm{src}}}{2dS_{\mathrm{src}}}
    \right)
    =
    100\left(\frac{d-r}{2d}\right).
    \label{eq:app_sliced_net_change}
\end{equation}
This is the main efficiency advantage of slicing. \PLRAttnRes and \LRAttnRes have the same low-rank depth-wise kernel, but only \LRAttnRes removes the projected-key path entirely.

\subsection{Activation memory}

Let $K_{\mathrm{src}}$ be the number of retained depth sources. For full \AttnRes, this is the number of previous residual-writing sources. For \BlockAttnRes, this is the number of retained block sources. Assume a fused depth-wise attention kernel, analogous to FlashAttention, so the attention-weight vector is not materialized for backward. If each activation element uses $b$ bytes, the dominant source-cache activation memory for ordinary \AttnRes is
\begin{equation}
    A_{\mathrm{AttnRes}}
    =
    bK_{\mathrm{src}}d.
    \label{eq:app_attnres_activation}
\end{equation}

\PLRAttnRes caches the full-dimensional values and the projected low-rank keys:
\begin{equation}
    A_{\mathrm{P}}
    =
    bK_{\mathrm{src}}(d+r).
    \label{eq:app_projected_activation}
\end{equation}
Therefore the retained source-cache activation memory increases by
\begin{equation}
    \%\,\Delta A_{\mathrm{P}}
    =
    100\left(
    \frac{A_{\mathrm{P}}}{A_{\mathrm{AttnRes}}}-1
    \right)
    =
    100\frac{r}{d}.
    \label{eq:app_projected_activation_increase}
\end{equation}

\LRAttnRes can implement the key as a view into the last $r$ coordinates of the cached value. In that implementation, it does not require a separate key cache:
\begin{equation}
    A_{\mathrm{S}}
    =
    bK_{\mathrm{src}}d
    =
    A_{\mathrm{AttnRes}}.
    \label{eq:app_sliced_activation}
\end{equation}
Thus the dominant retained source-cache activation increase is
\begin{equation}
    \%\,\Delta A_{\mathrm{S}}
    =
    0.
    \label{eq:app_sliced_activation_increase}
\end{equation}
An implementation may optionally cache normalized sliced keys, but this is not required by the method.

\subsection{Parameters}

The parameter count for \PLRAttnRes consists of projected key rows and residual-site queries. For residual-writing sub-layer $i$, adding an $r$-dimensional key to an output projection with input width $m_i$ adds $rm_i$ parameters. The embedding key projection adds $rd$ parameters. The $2L$ query-bearing read sites each have one $r$-dimensional static query, adding $2Lr$ parameters. Thus
\begin{align}
    \Delta P_{\mathrm{P,key}}
    &=
    r\sum_{i=1}^{2L}m_i + rd, \\
    \Delta P_{\mathrm{P,query}}
    &=
    2Lr, \\
    \Delta P_{\mathrm{P}}
    &=
    r\sum_{i=1}^{2L}m_i + rd + 2Lr.
\end{align}
With one attention output projection of width $d$ and one SwiGLU output projection of width $m$ per Transformer block,
\begin{equation}
    \Delta P_{\mathrm{P}}
    =
    r\bigl[L(d+m)+d+2L\bigr].
    \label{eq:app_projected_param_delta}
\end{equation}

\LRAttnRes has no key-projection parameters. It only needs the low-rank residual-site queries:
\begin{equation}
    \Delta P_{\mathrm{S}}
    =
    2Lr.
    \label{eq:app_sliced_param_delta}
\end{equation}

Relative to a base model with $P_{\mathrm{base}}$ parameters, the parameter increases are
\begin{align}
    \%\,\Delta P_{\mathrm{P}}
    &=
    100
    \frac{
    r\bigl[L(d+m)+d+2L\bigr]
    }{
    P_{\mathrm{base}}
    },
    \label{eq:app_projected_param_percent}\\
    \%\,\Delta P_{\mathrm{S}}
    &=
    100
    \frac{2Lr}{P_{\mathrm{base}}}.
    \label{eq:app_sliced_param_percent}
\end{align}

If ordinary \AttnRes is the reference instead of the no-\AttnRes baseline, then ordinary \AttnRes already contains $2Ld$ query parameters under the same parameter-free RMSNorm convention. The parameter differences become
\begin{align}
    \Delta P_{\mathrm{P-vs-AttnRes}}
    &=
    r\bigl[L(d+m)+d+2L\bigr] - 2Ld,
    \label{eq:app_projected_vs_attnres_params}\\
    \Delta P_{\mathrm{S-vs-AttnRes}}
    &=
    2Lr - 2Ld
    =
    -2L(d-r).
    \label{eq:app_sliced_vs_attnres_params}
\end{align}
Thus sliced low-rank routing uses fewer query parameters than full-dimensional \AttnRes and adds no key-projection parameters.

\subsection{Numerical example for the experimental model}

For an illustrative $r=32$ configuration of the $0.5\text{B}$ model, $d=1024$, $L=24$, and $m=2816$. The low-rank depth-wise residual kernel reduction is
\begin{equation}
    100\left(\frac{1024-32}{2\cdot 1024}\right)
    =
    48.44\%.
\end{equation}
Both \PLRAttnRes and \LRAttnRes obtain this kernel reduction.

The projected-key path in \PLRAttnRes adds
\begin{equation}
    C_{\mathrm{P,key\text{-}proj}}
    =
    32\left[24(1024+2816)+1024\right]
    =
    2{,}981{,}888
\end{equation}
multiply-adds per token. \LRAttnRes removes this term.

The parameter increase for \PLRAttnRes{} is
\begin{equation}
    \Delta P_{\mathrm{P}}
    =
    32\left[24(1024+2816)+1024+48\right]
    =
    2{,}983{,}424,
\end{equation}
while the sliced-key parameter increase is only
\begin{equation}
    \Delta P_{\mathrm{S}}
    =
    2\cdot 24\cdot 32
    =
    1{,}536.
\end{equation}

For retained source-cache activations, \PLRAttnRes increases the value-plus-key cache by
\begin{equation}
    100\frac{32}{1024}
    =
    3.125\%,
\end{equation}
whereas \LRAttnRes adds no separate source-key cache when implemented as a slice of the cached value.

For a dense SwiGLU Transformer core with
\begin{equation}
    C_{\mathrm{base}}
    =
    L(4d^2 + 3dm),
\end{equation}
the projected-key parameter overhead, ignoring embeddings and output heads, is approximately
\begin{equation}
    100\frac{32}{1024}
    \frac{1+2816/1024}{4+3(2816/1024)}
    =
    0.957\%.
\end{equation}
The exact projected-key numerator above gives roughly the same scale, while the sliced-key query-only overhead is negligible.

Under the added-FLOP accounting used in Table~\ref{tab:validation_results_all}, the projected-key path contributes approximately $0.9673\%$ of the non-embedding Transformer core for $r=32$. Thus the full projected $r=32$ row has total added FLOPs $0.4193\% + 0.9673\% \approx 1.387\%$, while block projected $N=8,r=32$ has total added FLOPs $0.09043\% + 0.9673\% \approx 1.058\%$. \LRAttnRes{} has no projected-key path, so its total added FLOPs equal its depth-kernel compute.

\newpage
\begingroup
\setlength{\intextsep}{6pt plus 2pt minus 2pt}
\section{1B and 4B evaluation details}
\label{app:preliminary_1b}

\paragraph{Downstream protocol.}
We use \texttt{lm-evaluation-harness} v0.4.7 with its task prompts, zero demonstrations, the \texttt{cl100k\_base} tokenizer, and a 2048-token context. A custom adapter scores answer continuations with BF16 model computation and FP32 log-probabilities, using batches of eight and causal right padding. Prompt and answer tokenization preserves boundary whitespace. Overlong inputs are left-truncated while retaining the complete answer, and an end-of-text prefix is used only for an empty prompt. There is no chat template or additional beginning-of-sequence token.

We evaluate the complete test splits of ARC-Challenge (1172 items), ARC-Easy (2376), and OpenBookQA (500), and the validation splits of BoolQ (3270), CommonsenseQA (1221), HellaSwag (10{,}042), PIQA (1838), and Social-IQA (1954). We report \texttt{acc\_norm} for ARC-Challenge, ARC-Easy, HellaSwag, OpenBookQA, and PIQA, selecting the answer with the highest log-likelihood divided by its character length. BoolQ, CommonsenseQA, and Social-IQA use \texttt{acc}, based on unnormalized answer log-likelihood. The reported average weights the eight task accuracies equally.

\begin{table}[!ht]
    \centering
    \caption{Zero-shot downstream accuracy (\%) for the $1\text{B}$ and $4\text{B}$ pairs. Metrics and evaluation splits are specified above. Bold marks the higher accuracy within each scale.}
    \label{tab:downstream_1b}
    \vspace{0.65em}
    \small
    \setlength{\tabcolsep}{5pt}
    \newcommand{\DownstreamResultsRows}{%
ARC-Challenge & \textbf{28.58} & 28.24 & 28.07 & \textbf{28.67} \\
ARC-Easy & 52.06 & \textbf{52.10} & 53.49 & \textbf{55.39} \\
BoolQ & \textbf{59.79} & 57.89 & 61.25 & \textbf{61.41} \\
CommonsenseQA & 18.84 & \textbf{20.39} & 20.39 & \textbf{20.97} \\
HellaSwag & 39.23 & \textbf{40.13} & 41.10 & \textbf{42.38} \\
OpenBookQA & 31.20 & \textbf{32.20} & \textbf{33.20} & 31.40 \\
PIQA & 67.74 & \textbf{68.06} & 68.55 & \textbf{68.72} \\
Social-IQA & 37.51 & \textbf{39.10} & 37.72 & \textbf{39.41} \\
\midrule
Eight-task mean & 41.87 & \textbf{42.27} & 42.97 & \textbf{43.54} \\
}
    \begin{tabular}{@{}lrrrr@{}}
        \toprule
        & \multicolumn{2}{c}{$1\text{B}$} & \multicolumn{2}{c}{$4\text{B}$} \\
        \cmidrule(lr){2-3}\cmidrule(lr){4-5}
        Task & \AttnRes{} & \LRAttnRes{} & \AttnRes{} & \LRAttnRes{} \\
        \midrule
        \DownstreamResultsRows
        \bottomrule
    \end{tabular}
\end{table}

\begin{figure}[H]
    \centering
    \includegraphics[width=\linewidth]{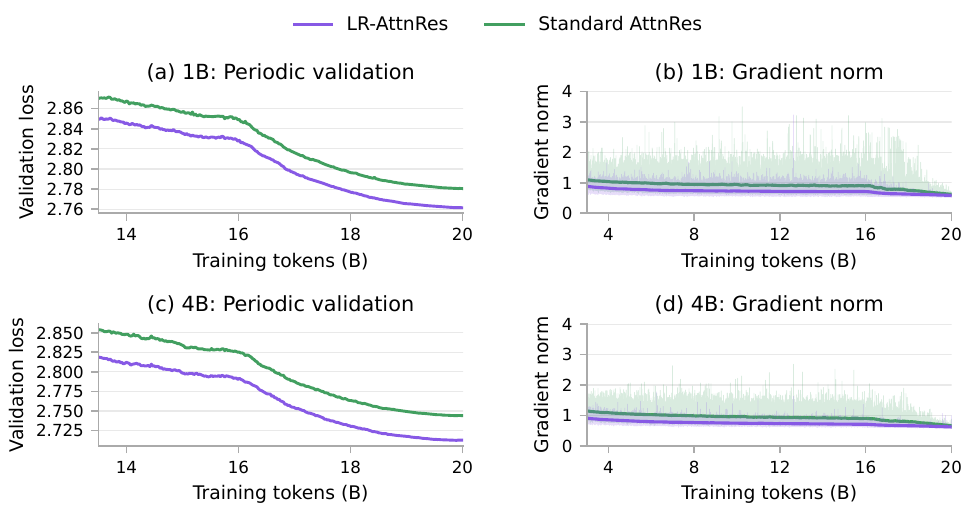}
    \caption{Training dynamics at $1\text{B}$ (top) and $4\text{B}$ (bottom), with $r=d/4$ for \LRAttnRes{}. Left: periodic validation loss over the final $6.5\text{B}$ training tokens. Right: pre-clipping gradient norms from $3\text{B}$ to $20\text{B}$ tokens on linear axes, capped at 4 for display. Faint traces are individual steps and solid curves are trailing 1,000-step means.}
    \label{fig:preliminary_1b}
\end{figure}

\subsection{Gradient-norm statistics}
\label{app:large_model_gradients}

Table~\ref{tab:large_model_gradients} summarizes pre-clipping gradient norms over all $76{,}294$ training steps. Across the full training runs, \LRAttnRes{} reduces mean gradient norm by $19.5\%$ at $1\text{B}$ and $21.2\%$ at $4\text{B}$, and population variance by $30.8\%$ and $34.1\%$. The fraction of logged norms above the clipping threshold of 1 falls from $35.1\%$ to $11.0\%$ and from $38.3\%$ to $9.9\%$, respectively.

\begin{table}[!ht]
    \centering
    \caption{Full-run gradient-norm statistics at $1\text{B}$ and $4\text{B}$. The final column is the percentage of steps with pre-clipping norm above 1.}
    \label{tab:large_model_gradients}
    \vspace{0.65em}
    \small
    \setlength{\tabcolsep}{6pt}
    \begin{tabular}{@{}llrrr@{}}
        \toprule
        Model & Method & Mean & Variance & Above 1 (\%) \\
        \midrule
        1B & \AttnRes{} & 0.9754 & 0.1747 & 35.1 \\
        1B & \LRAttnRes{} & 0.7853 & 0.1209 & 11.0 \\
        \midrule
        4B & \AttnRes{} & 1.022 & 0.1679 & 38.3 \\
        4B & \LRAttnRes{} & 0.8055 & 0.1106 & 9.9 \\
        \bottomrule
    \end{tabular}
\end{table}

\section{PyTorch-style pseudocode}
\label{app:lr_attnres_pseudocode}

\begin{figure}[H]
\begin{lstlisting}[style=lrattnres]
def route(sources, q, r):
    V = torch.stack(sources)                   # [S, B, T, d]
    K = rms_norm(V[..., -r:])                  # [S, B, T, r]
    weights = (K * q).sum(-1).softmax(dim=0)
    return (weights[..., None] * V).sum(dim=0)

def lr_attnres(x, sublayers, queries, r, N=None):
    M = len(sublayers)                         # M = 2L
    N = M if N is None else min(N, M)
    ends = {ceil(M * j / N) for j in range(1, N + 1)}
    blocks, partial, h = [x], None, x

    for t, (f, q) in enumerate(zip(sublayers, queries), start=1):
        y = f(rms_norm(h))                     # attention or SwiGLU
        partial = y if partial is None else partial + y
        if t in ends:
            blocks.append(partial)
            partial = None
        sources = blocks + ([] if partial is None else [partial])
        h = route(sources, q, r)
    return rms_norm(h)                         # then the LM head
\end{lstlisting}
\caption{PyTorch-style pseudocode for full and block \LRAttnRes. \texttt{x} is the token embedding. \texttt{sublayers} alternates attention and SwiGLU transformations without residual addition. The $2L$ static $r$-dimensional queries are learned and initialized to zero. \texttt{rms\_norm} normalizes the last dimension. \texttt{N=None} retains individual outputs (full routing), whereas a specified $N$ groups them into block sums. Setting $r=d$ recovers standard \AttnRes with the same source grouping.}
\label{fig:lr_attnres_pseudocode}
\end{figure}

\endgroup
\section{Fast-AttnRes kernels and training throughput}
\label{app:fast_attnres}

\paragraph{Implementation.}
\fastattnres{} supports standard routing ($r=d$) and sliced routing ($r<d$) through a shared source-list interface. Building on the source-pointer and tiled online-softmax design of FLA~\citep{yang2024fla}, our implementation reads the last $r$ coordinates of each source for parameter-free RMS normalization and query scoring, while retaining all $d$ coordinates for the weighted value sum. The same operator serves full and block routing. The model determines which sources are retained. BF16 inputs and outputs use FP32 internal accumulation. The backward pass combines the direct value gradient with the routing gradient on the selected coordinates before writing the BF16 source gradient, and separately reduces the query gradient. Tiling supports unaligned widths and ranks. The source-list path avoids stacking values, saves normalization and softmax statistics, and recomputes the mixed value only for lists of up to three sources. PyTorch custom operators provide autograd and \texttt{torch.compile} integration. The implementation and its FLA attribution are available in the versioned source release.

\paragraph{Numerical correctness.}
We test the trained $1$B and $4$B checkpoints on H100 using one 2048-token batch per case. At each routing read, the native operator and independent PyTorch/autograd reference receive identical BF16 inputs and loss-derived upstream gradients. All 408 reads across six cases pass the elementwise BF16 criterion $|x-x_{\mathrm{ref}}|\leq 0.05+0.05|x_{\mathrm{ref}}|$. Table~\ref{tab:fast_attnres_correctness} shows comparable relative errors for FLA and \fastattnres{} at full width. Reduced-rank checks cover the non-power-of-two ranks $384$ and $640$.

\begin{table}[H]
    \centering
    \caption{Maximum relative $\ell_2$ errors (\%) against PyTorch on trained checkpoints. \textbf{All cases pass the BF16 tolerance}. Maxima cover 48 reads at $1$B, 88 at $4$B, and independent source gradients. FLA uses checkpoint level 1.}
    \label{tab:fast_attnres_correctness}
    \vspace{0.65em}
    \small
    \newcommand{\TrainedCorrectnessRows}{%
1B & FLA & 1536 & 0.003309 & 0.29 & 0.02017 \\
1B & \fastattnres{} & 1536 & 0.004766 & 0.2899 & 0.01689 \\
1B & \fastattnres{} & 384 & 0.002804 & 0.2179 & 0.01557 \\
\midrule
4B & FLA & 2560 & 0.003682 & 0.3029 & 0.01388 \\
4B & \fastattnres{} & 2560 & 0.003682 & 0.3007 & 0.0127 \\
4B & \fastattnres{} & 640 & 0.003156 & 0.2636 & 0.03336 \\
}
    \begin{tabular}{@{}llrrrr@{}}
        \toprule
        Model & Implementation & $r$ & Forward & Source grad. & Query grad. \\
        \midrule
        \TrainedCorrectnessRows
        \bottomrule
    \end{tabular}
\end{table}

Under the timing software stack below, B200 checks of all three configurations at both model widths and source counts $2,5,9,10$ pass the same criterion for outputs and both gradients (24 cases with 16 random token rows each). Separate B200 checks with PyTorch 2.13.0 and Triton 3.7.1 cover unaligned width $d=513$, repeated or strided sources, strided queries, and changed-input replays. All pass the same criterion.

\paragraph{Matched model comparison.}
All timings use $N=8$, with FLA at $r=d$ and \fastattnres{} at $r=d$ or $d/4$. Methods share the production Transformer, document masking, vocabulary head, training objective, and Muon/Adam optimizers. The $1$B and $4$B models use $(L,d,m)=(24,1536,4224)$ and $(44,2560,7040)$, respectively, with head width 64, context length 2048, and vocabulary size $100{,}277$.

\paragraph{Measurement protocol.}
Each three-method repeat uses matched shared weights, seed, cached real training tokens, and production document boundaries. Both GPUs use BF16, PyTorch 2.10.0 (CUDA 13.0), Triton 3.6.0, FlashAttention 2.8.3, and \texttt{fullgraph=True, dynamic=False}, without CUDA Graph capture. Microbatch size 2 and 64 accumulation steps give $262{,}144$ tokens per update. Synchronized timing includes host-to-device transfer, forward, backward, accumulation, clipping, and optimizer updates. Compilation, storage reads, evaluation, and logging are excluded. Three warmup updates precede eight to thirty measured updates. Timing stops when two consecutive six-update windows each have a coefficient of variation of at most $1\%$ and drift of at most $0.5\%$ between their halves. All measured updates contribute. Each GPU uses five fresh repeats per configuration (30 measurements), with rotated method order and matched production code, data, and shared backbone initialization across GPUs.

\paragraph{Paired uncertainty.}
For each of the five matched repeats, let $v_j^{(d)}$ and $v_j^{(d/4)}$ denote complete training-step throughput under \fastattnres{}. We compute the percentage gain and its approximate $95\%$ Student-$t$ confidence interval as
\[
    g_j=100\left(\frac{v_j^{(d/4)}}{v_j^{(d)}}-1\right),
    \qquad \bar g \pm t_{0.975,4}\frac{s_g}{\sqrt{5}},
\]
where $s_g$ is the sample standard deviation of the five gains. Each observation is a matched pair of fresh processes. The interval assumes independent, approximately normally distributed repeat-level gains. Training-throughput gains from changing the kernel at fixed rank, computed from mean throughput at $r=d$, are $0.95\%$ and $0.74\%$ at $1\text{B}$ and $4\text{B}$ on H100, and $2.17\%$ and $1.75\%$ on B200.

\end{document}